\documentclass[lettersize,journal]{IEEEtran}
\usepackage{amsmath,amsfonts}
\usepackage{algorithmic}
\usepackage{algorithm}
\usepackage{array}
\usepackage[caption=false,font=normalsize,labelfont=sf,textfont=sf]{subfig}
\usepackage{textcomp}
\usepackage{stfloats}
\usepackage{url}
\usepackage{verbatim}
\usepackage{graphicx}
\usepackage{cite}
\usepackage{makecell}
\usepackage[table]{xcolor}
\usepackage{nicefrac}       
\usepackage{microtype}      
\usepackage{siunitx} 
\usepackage{cite}
\usepackage{amsmath,amssymb}
\usepackage{algorithmic}

\usepackage{array}
\usepackage{booktabs}
\makeatletter
\def\thickhline{%
  \noalign{\ifnum0=`}\fi\hrule \@height \thickarrayrulewidth \futurelet
   \reserved@a\@xthickhline}
\def\@xthickhline{\ifx\reserved@a\thickhline
               \vskip\doublerulesep
               \vskip-\thickarrayrulewidth
             \fi
      \ifnum0=`{\fi}}
\makeatother

\newlength{\thickarrayrulewidth}
\begin{document}
\title{A Compliant Robotic Leg Based on \\Fibre Jamming}

\author{Lois Liow, James Brett, Josh Pinskier, Lauren Hanson, Louis Tidswell, Navinda Kottege ~\IEEEmembership{Senior Member,~IEEE} and David Howard ~\IEEEmembership{Member,~IEEE}

\thanks{L. Liow, J. Brett, J. Pinskier, L. Hanson, N. Kottege, and D. Howard are with the Robotics and Autonomous Systems Group, CSIRO, Pullenvale, QLD 4069, Australia. All correspondence should be addressed to {\tt\small lois.liow@csiro.au}}

\thanks{L. Tidswell is with the Queensland University of Technology (QUT), Gardens Point, QLD 4000, Australia and the Robotics and Autonomous Systems Group, CSIRO, Pullenvale, QLD 4069, Australia}

\thanks{Manuscript received July 2023}}

\markboth{IEEE Transactions on Robotics}%
{Liow \MakeLowercase{\textit{et al.}}: A Compliant Robotic Leg Based on
Fibre Jamming}


\maketitle

\begin{abstract}
Humans possess a remarkable ability to react to unpredictable perturbations through immediate mechanical responses, which harness the visco-elastic properties of muscles to maintain balance.  Inspired by this behaviour, we propose a novel design of a robotic leg utilising fibre jammed structures as passive compliant mechanisms to achieve variable joint stiffness and damping. We developed multi-material fibre jammed tendons with tunable mechanical properties, which can be 3D printed in one-go without need for assembly. Through extensive numerical simulations and experimentation, we demonstrate the usefulness of these tendons for shock absorbance and maintaining joint stability. We investigate how they could be used effectively in a multi-joint robotic leg by evaluating the relative contribution of each tendon to the overall stiffness of the leg. Further, we showcase the potential of these jammed structures for legged locomotion, highlighting how morphological properties of the tendons can be used to enhance stability in robotic legs.

\end{abstract}


\section{Introduction}
\IEEEPARstart{L}{egged} robotics is a rapidly evolving field, with recent applications ranging from search-and-rescue operations \cite{spenko2018darpa} to the exploration of complex and unstructured environments\cite{hutter2016anymal,catalano2021adaptive}. However, despite significant advances in the research of legged robotics, developing a closed-loop control system that involves feedback from sensors to actively control the trajectory of the leg remains a challenge for researchers. This is largely due to the unpredictability and varying nature of real-world terrains, which cannot be wholly accounted for or pre-planned in the control design of a robotic system. Hence, there is a push for robots to be able to handle unexpected events with minimal sensory feedback and control. One approach to addressing these challenges is to implement morphologically intelligent structures, which allow robotic legs to passively adapt to changing terrains and unexpected perturbations through its material makeup. An example of this adaptive behaviour can be seen in humans, whereby they are able to perform "preflexes", an auto-corrective movement response, provided by the passive tension of the muscles. Preflexes play an important role in the dynamic stability of both humans and animals, and this is enabled by the intrinsic mechanical properties in the muscles \cite{loeb1995control,moritz2004passive}. Muscles, are visco-elastic in nature and are able to passively perform perturbation rejection almost instantly when there is a change in length\cite{izzi2023muscle}. This is especially useful in unforeseeable situations such as falling into a hole or tripping on a curb, where the foot in unexpectedly stretched up or down.   

\begin{figure}[!t]
\centerline{\includegraphics[width=\columnwidth]{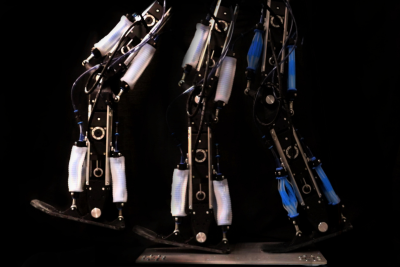}}
\caption{Our Jamming lEG, JEG, is a belt-driven robotic leg with antagonistic variable stiffness multi-material fibre jammed tendons.}
\label{fig:JEG}
\end{figure}

Robotic legs are typically comprised of rigid elements connected by stiff joints. However, in reality, nature's design tends to favour more soft materials and yet, they are able to perform robustly in the harshest of environments. Even stiff structures such as bones, are surrounded by soft tissues like muscles and tendons. While rigid structures afford more precise control and possess greater load-bearing and force transmission capabilities, soft structures enable passive adaptation to the environment. The flexibility inherent in soft robotics presents a cost-effective alternative for researchers, as it eliminates the need for high-frequency feedback control systems that would otherwise be necessary to prevent damage to the robot. Hence, this resulted in the emergence of articulated soft systems, which combines the controllability of rigid structures with the flexibility of compliant structures \cite{della2020soft}. Researchers have long exploited this paradigm to create robotic systems that can better adapt and perform in various environments. By bridging the gap between rigid and soft robotics, we open up new possibilities for more morphologically intelligent and versatile designs.


Compliance in legged robots can be achieved through various techniques, including Pneumatic Artificial Muscles (PAM)\cite{kalita2022review,takuma20083d,rosendo2015stretch,mirvakili2020actuation,lei2017joint}, Series Elastic Actuators (SEA) \cite{vanderborght2011maccepa,guenther2019improving} and by integrating passive elastic elements into the leg's mechanism \cite{fedorov2017design,sato2017design,abad2019significance}. PAMs have been commonly used by researchers in the past due to their high power-to-weight ratio \cite{kalita2022review}. However, the requirement for an air compressor, which are often heavy and lack portability, often limits their applications. Furthermore, the non-linearities within the structure presents a perennial challenge to researchers trying to control this dynamic behaviour via pressure \cite{kalita2022review}. Elastic elements or mechanical springs, on the other hand, are a much more popular choice due to their ease of implementation and modelling. However, these springs have limited tunability due to their fixed linear stiffness and consequently, are unlikely to be able to handle a variety of different loading conditions, as one would experience when traversing diverse natural environments. Furthermore, selection of these springs were traditionally based on designers intuition or by estimations from  experimental human locomotion data \cite{radkhah2011concept}. To solve this problem, Variable Stiffness Actuators (VSAs) had been introduced \cite{wolf2015variable}, which requires two motors to control both the equilibrium position and stiffness of the joint. However this advantage comes with a drawback, as VSAs have greater mechanical complexity and add more weight to the robotic system compared to traditional SEAs \cite{rodriguez2019variable,deboer2022discrete}. In more recent works, Non-linear Series Elastic Actuators \cite{deboer2022discrete, qian2022toward,okken2022progressive} have been proposed as alternatives to VSAs to create progressive non-linear force-deflection similar to that found in human muscles \cite{okken2022progressive}. While these technologies have made significant advances, their design complexity pose greater challenges and costs in manufacturing, making them largely out of reach for researchers. In the recent years, there has been a major surge in the research surrounding soft materials and structures, which challenges the use of traditional robotic systems. Elastomer-based Series Elastic Actuators (eSEA) have been used in place of traditional spring-based systems, due to their inherent damping properties \cite{jarrett2019modeling, paskarbeit2013self}.

In this work we focus particularly on jamming \cite{fitzgerald2020review,manti2016stiffening}, and investigate how variable stiffness and damping in robotic legs could be achieved. Jamming, although a popular variable stiffness technology used in soft robotics, has to date been relatively understudied for legged applications. Due to the potential for rapid, high-range stiffness variation, jammed structures have shown promise in related works, e.g., to achieve variable stiffness in wearables, such as exoskletons \cite{shen2020scalable}, as a friction based locking mechanisms for revolute joints\cite{sozer2022novel}, and flexure mechanisms for robot wrists \cite{aktacs2019flexure}. Ease of manufacturability and cost had been the primary drivers for the widespread adoption of this technology, with major work done in streamlining rapid prototyping processes, such as 3D printing, to fabricate these structures in 'one-shot', whereby the membrane and grains of entire grippers were printed in a single print run\cite{howard2022one}.

There are three main jamming technologies available to us; granular, layer, and fibre jamming.  Granular jamming, which consists of grains sealed in an air-tight membrane, is by far the most popular method used to achieve a variable stiffness structure\cite{fitzgerald2020review}. Under compression, a large variation in stiffness can be achieved and hence, researchers in the past have exploited this mechanical property to improve foot adaptability in various terrains \cite{lathrop2020shear,hauser2016friction,chopra2020granular}. However, these adaptive feet are often made from delicate soft materials, such as latex or textile membranes filled with granular material, which are prone to wear in rough environments. Furthermore, when a torsional or bending force is applied, grains tend to slip and can result in unfavourable configurations, making it less appropriate to be used in revolute joints, as it may inhibit their motion. The non-homogeneous distribution of grains can lead to variable mechanical properties across the structure and consequently, making it difficult for researchers to quantify the structural stiffness.

As well as granular jamming, recent years have seen developments in alternative jamming mechanisms including layer and fibre jamming. Layer jamming structures are by far the least deformable \cite{fitzgerald2020review} and their stiffness is highly dependent on planar orientation of the layers. This limits the number of DOF on a joint and any motion that is out of plane may introduce unwanted stiffness or result in buckling, which is undesirable in legged locomotion. 

Fibre jamming, on the other hand, offers a low bending stiffness in the two orthogonal directions to the axis of the fibres (the radial direction), which is ideal to allow freedom-of-movement of the rotational joints in the leg \cite{yang2021reprogrammable}. Fibre Jammed Structures (FJS) utilise bundles of longitudinal fibres fixed at both ends in a system enclosed within a sealed membrane\cite{brancadoro2018preliminary,aktacs2021modeling,yang2021reprogrammable,jadhav2022variable,arleo2023variable}. These flexible fibres are free to slide across each other and thus, allowing low bending stiffness in all directions\cite{aktacs2021modeling}, which cannot be achieved by traditional granular and layer jamming methods. In their unjammed state, these fibres are free to rearrange against each other within the membrane. However, once vacuum pressure is applied, these fibres are constricted by the membrane and the structure becomes rigid.

To the authors' knowledge, the implementation of fibre jammed structures to achieve joint compliance variability in legged robots have not thoroughly been investigated in literature, and previous focus was mainly in soft jammed feet \cite{howard2022one,lathrop2020shear,hauser2016friction,chopra2020granular}. This is largely due to the complexity in understanding the behaviour of these soft structures under constantly changing loading conditions, such as walking. To address this problem, we introduce a novel robotic leg, JEG (Jammed Leg) which utilises fibre jammed tendons in series with the belt transmission system to provide passive joint compliance and damping to prevent damage to the robot during collisions (Fig. \ref{fig:JEG}). We mainly focus on introducing fibre jammed structures to mainly to the knee and ankle, as they are known to be the primary shock absorbers during rapid loading of the leg\cite{shen2022neuromechanical,derrick1998energy}.

The rest of the paper is organised as follows: Section II describes the design and fabrication of the multi-material fibre jammed tendons, along with tensile experiments to characterise the stiffness and damping properties of different tendon design configurations, including tendons with varying fibre diameters and number of fibre layers. We then propose an improved FEM numerical simulation model of the tendon, which is able to capture the frictional interactions between tendon fibres during the jamming transition. By doing so, the model provides us with a comprehensive understanding of the mechanism underlying the distinct mechanical properties observed in the unjammed and jammed states of the tendons. Section III, demonstrates the performance of the tendons in an antagonistic lever-type mechanism to vary the impact response of a revolute joint. Section IV discusses the design of the fibre jammed leg---JEG, by first introducing the mechanical design of the robotic leg, followed by the belt driven linear actuator design with details on the placement of the tendons. In Section V, a multibody FEM simulation was used to simulate behaviour of the JEG and the tendons upon collision with an obstacle, which resulted in either a toe-down or toe-up adaptation. We evaluate the response of the JEG towards obstacle-induced perturbation by analysing the changes in forces and energy of the system prior and after the 
collision. This is followed by Section VI, which evaluates the role and force contribution of each tendon pair towards the overall stiffness of the JEG in response to rotational perturbations. Section VII assesses the walking performance of the JEG empirically by analysing the Ground Reaction Forces (GRF) for different jamming patterns using a force plate. In Section VIII, we explore and discuss results obtained from the simulations and experiments performed. Finally, we conclude the paper in Section IX with proposed future work and a discussion on limitations in Section X.

Our previous work \cite{pinskier2022jammkle} introduced the  3D printed fibre jammed tendon unit, together with numerical and analytical methods to analyse its stiffness variability. In this paper, we significantly expand this fundamental work by (I) designing and fabricating a bespoke articulated soft robot leg that harnesses antagonistic tendon pairs for stable walking, (II) performed a full characterisation of a multitude of aspects of performance in both FEA and experimentation, and (III) showed the potential of such systems to be used to enhance disturbance rejection and stability of a multi-joint robotic system.

\begin{figure*}[t!]
\centerline{\includegraphics[width=1.95\columnwidth]{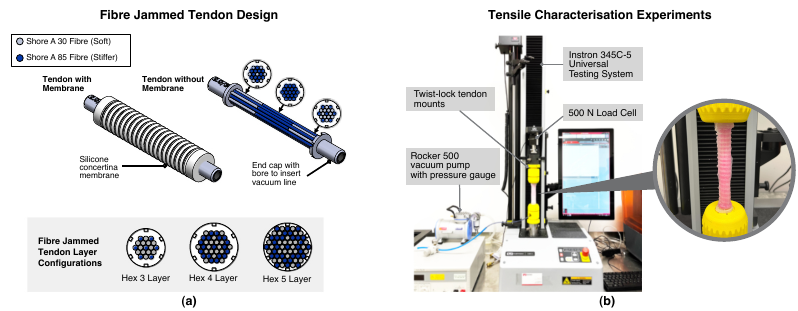}}
\caption{ (a) Left: Fibre jammed tendon with concertina membrane. Right: Fibre jammed tendon with fibre layout shown for different sections along the length of the tendon.  Multi-material fibres consists of shorter sections of soft shore A 30 (clear segment) and longer sections of stiff shore A 85 (blue segment) fibres. (b) Experimental setup to perform tensile tests, where tendons are are stretched up to 20\,mm in their unjammed and jammed states.}
\label{fig:tendon}
\end{figure*}

\section{Design and Fabrication of Fibre Jammed Tendons}

\subsection{Tendon Design}

Fibre jamming was selected over layer jamming due to their tensile stiffness tunability and ease of integration into  antagonistic pull-rod type joint systems. The longitudinal tension provided by the fibre jammed tendons enable them to be used like extension springs and dampers, which have been used traditionally for these antagonistic type mechanisms. Layer jamming in contrast, can only be tuned for bending stiffness \cite{yang2021reprogrammable}, and hence is unsuitable for this application.

The design of the multi-material fibre jammed tendons are shown in Fig. \ref{fig:tendon} (a). The tendons were fabricated using a Stratasys Connex3 Objet500 PolyJet 3D printer with a combination of Vero (rigid Shore D) and Agilus30 (flexible Shore A 30) material. By selecting different proportions of these resins as a mixture, composite materials of various Shore A hardness could be produced. The fibres were fixed together at each end in a Fixed-Bundle Type (FBT) configuration with pure Vero rigid end plates. The overall length of the tendons including the end caps is 120\,mm, while the fibres are 80\,mm length with material distribution ratio of two-third Shore A 85 (55\,mm) and one-third Shore A 30 (30\,mm). Tendons of varying number of fibre layers (3,4, and 5) and fibre diameter (1.5\,mm, 2.0\,mm, 2.5\,mm, and 3.0\,mm) were produced. One end of the end plates was designed with a bore that attaches to a vacuum line. The use of a multi-material printer is advantageous because these end plates and fibres can be printed together as a single part. Hence, no assembly or handling of these intricate fibres were required.

The fibres were placed in a hexagonal arrangement for higher packing density in comparison to a round and square packing configuration \cite{pinskier2022jammkle}. A hexagonal arrangement enabled a greater area of contact between the surface of the fibres and consequently produces a higher jamming force\cite{pinskier2022jammkle}. To maximise contact area between different materials (Shore A 30 and A 85), inverted fibre layouts were used at the opposite ends of the tendons, as seen in Fig. \ref{fig:tendon} (a). A silicone rubber (Smooth-on Ecoflex 00-30) cast concertina membrane was then stretched over the fibre bundle to create an enclosed vacuum space.  When negative pressure was applied, the membrane forces the fibres together and create a jamming effect. 

\begin{figure*}[t!]
\centerline{\includegraphics[width=1.95\columnwidth]{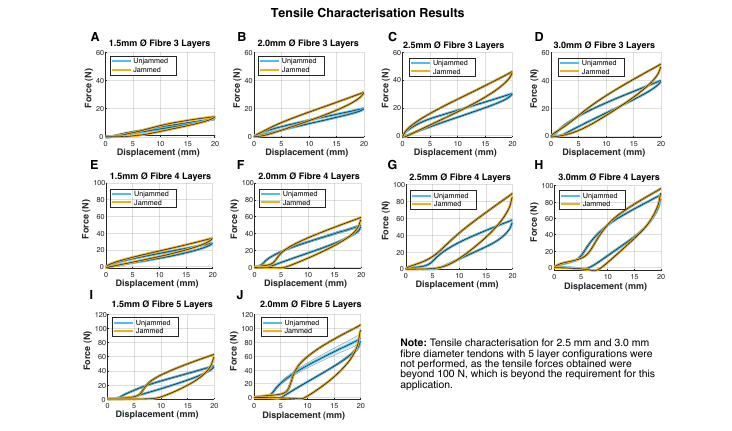}}
\caption{(A-J) Plots showing tensile force vs displacement for fibre jammed tendons of 1.5, 2.0, 2.5, and 3.0\,mm fibre diameters in hexagonal 3, 4, and 5 layer configurations, while unjammed (0\,kPA) and jammed (50\,kPA). }
\label{fig:tendonresults}
\end{figure*}

 \begin{table*}[t!]
\centering
\caption{\label{tab:tendondata} Tendon material properties obtained from tensile test.}
\begin{tabular}{p{1.2cm} p{0.8cm} p{2cm} p{1.75cm} p{1.2cm} p{1.8cm} p{1.8cm} p{1.7cm}}
\toprule
\textbf{Fibre Diameter (mm)} & \textbf{Layers} & \textbf{Mean Unjammed Force (SE) [N]} & \textbf{Mean Jammed Force (SE) [N]}  & \textbf{Stiffness Increase ($\%$)} & \textbf{Unjammed Damping Capacity} & \textbf{Jammed Damping Capacity} &
\textbf{Damping Capacity Increase ($\%$)} \\
\midrule
1.5 &  3 & 13.1 (1.51) & 16.0 (2.85) & 22 & 0.21 & 0.29 & 38\\
       &  4 &  23.6 (4.36) & 31.3 (2.74) & 32 & 0.28 & 0.33 & 18\\
       &  5 &  48.0 (2.58) & 68.0 (4.34) & 41 & 0.29 & 0.54 & 86\\
\midrule
2.0 &  3 & 20.5 (1.33) & 28.8 (2.04) & 40 & 0.20 & 0.36 & 80\\
       &  4 & 46.9 (0.90) & 57.8 (2.63) & 23 & 0.32 & 0.50 & 56\\
       &  5 & 85.7 (2.26) & 106.7 (2.84) & 25 & 0.39 & 0.66 & 69\\
\midrule
2.5 &  3 & 34.7 (3.37) & 46.2 (3.43) & 33 & 0.30 & 0.39 & 30\\
       &  4 & 67.4 (7.25) & 88.5 (10.24) & 31 & 0.37 & 0.51 & 38\\
\midrule
3.0 &  3 & 45.4 (4.18) & 64.7 (9.76) & 43 & 0.32 & 0.44 & 38\\
       &  4 & 88.4 (7.84) & 102.7 (10.35) & 16 & 0.45 & 0.52 & 16\\
\bottomrule

\end{tabular}

\end{table*}

\subsection{Tendon Stiffness and Damping Characterisation}

An Instron (Model 34SC-5 Single Column Table) universal testing machine was used to perform the tensile testing for tendons of various fibre diameters and layers. This helps us to tune the morphology of each tendon to its specific use case. For each tendon type, five samples were produced and the unjammed and jammed force-displacement data of these individual samples were collected using the experimental setup in Fig. \ref{fig:tendon} (b). The silicone concertina membranes were used during the entire tensile testing experiments, as our previous work has shown that it does not have an effect towards the measured stiffness \cite{pinskier2022jammkle}. Initially, no vacuum pressure was applied and each sample was stretched up to 20\,mm at a rate of 5\,mm/s before it was allowed to return to its relaxed state. This step was repeated for 5 cycles. Following this, the vacuum is then turned on to create a negative pressure of 50\,kPa, relative to the atmospheric pressure. The same sample was then stretched for another 5 cycles to collect the jammed force-displacement data, as plotted in Fig. \ref{fig:tendonresults}.

Results of our tendon characterisation tests for different fibre and layer configurations are shown in Table. \ref{tab:tendondata}. The mean unjammed and jammed forces were calculated by taking the average peak force from the force-displacement cycles. The fibre jammed tendons behave as Coloumb friction dampers to provide an adjustable impact response based on pressure, which will be further discussed in Section. \ref{section-damping}. Hysteresis is a fundamental behaviour seen in jammed structures, due to the increase in intrefibre contact area and hence, friction upon jamming. Hence, jamming increases not only stiffness of the fibre jammed tendon, but also its damping capacity, as seen in the force-displacement plots in Fig. \ref{fig:tendonresults}. The damping capacity (\textit{D}) of the tendons were calculated as the ratio of energy dissipated $\Delta U$ to energy stored ${U_{load}}$ in the cycling loading:

\begin{equation}
 D =\frac{\Delta U}{U_{load}}
\end{equation}

in which $\Delta U$ was calculated as the area between the loading-unloading curves and ${U_{load}}$ as area under the loading curve of the hysteresis loop.

\begin{figure*}[t]
\centerline{\includegraphics[width=1.6\columnwidth]{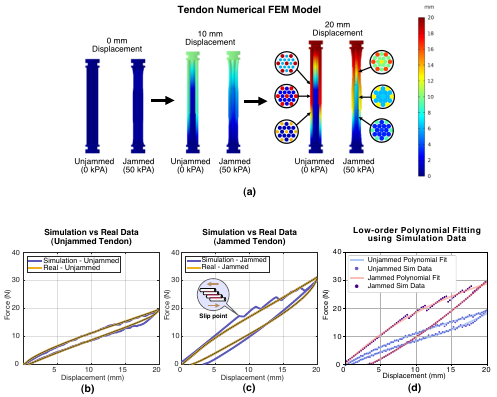}}
\caption{(a) FEM simulation of unjammed vs jammed (50\,kPa) tendon deformation with extension of up to
20\,mm. When unjammed, deformation is localised around the Shore A 30 (soft) sections of the individual fibres. However, upon jamming, the bundle of fibres deform as a single column and elongation is more distributed between the Shore A 30 (soft) and Shore A 85 (stiffer) materials. (b)(c)  Simulation model validation: Force-displacement curves of unjammed and jammed tendon upon elongation, comparing FEM simulation with real experimental results. Upon jamming, tensile stiffness and damping capacity of the tendon increases. Through simulation, the slip point, which is the onset at which frictional sliding occurs could be identified. Following the slip point, the fibres exhibit a stick-slip behaviour, as shown by the jagged region of the force-displacement curve. (d) Low-order polynomial fitting of simulation data to approximate the tendon's mechanical properties.}
\label{fig:fea}
\end{figure*}

From the results, the tensile force (and consequently, stiffness) and damping capacity of the tendons were seen to increase with fibre diameter and number of layers. The 1.5\,mm fibre diameter 5 Layers tendons showed the greatest increase in damping capacity upon jamming by 86\% and a significant increase in stiffness by 41\%. The 2.0\,mm fibre diameter 3 Layers tendons showed similar results with damping capacity and stiffness increase upon jamming of 80\% and 40\% respectively. From repeated experiments, we have observed that the fibres of the 1.5\,mm fibre diameter tendons were very intricate and hence, prone to wear quickly over time due to the current limitations of the 3D printing technology in fabricating thin features. From the results, tendons with larger fibre diameters (2.5\,mm and 3.0\,mm) were found to not offer much increase in damping capacity after jamming. This was mainly due to the large initial interfibre contact even while the tendons were unjammed. This could be mitigated in the future by increasing the spacing between neighbouring fibres to reduce initial contact between fibres.

\subsection{Tendon FEM Simulations}
\label{section-tendonfea}
To understand the mechanical behaviour of the fibre jammed tendon, the tendon was modelled using CAD and imported into COMSOL Multiphysics to perform FEM simulations, as shown in Fig. \ref{fig:fea} (a). A time-dependent study was set up with the same procedures used for the real tensile characterisation experiments. To reduce computational times, the tendon model was segmented and reduced to the size to 1/6th of its original model using the Symmetry tool in COMSOL. A fixed constraint was applied to the bottom end cap of the tendon, while a displacement (in the longitudinal direction of the fibres) was prescribed to the top end cap. An increasing displacement at a rate of 5\,mm/s was applied to the tendon until an extension of 20\,mm was reached, before it was allowed to return to its original length. Simulations were performed for both the tendon's unjammed and jammed states. As found in previous work \cite{pinskier2022jammkle}, the stiffness of the membrane is negligible, jamming is induced by applying a 50\,kPa pressure directly to the surface of the outermost tendons.

The material properties of the Shore A 30 and Shore A 85 printed materials were experimentally determined through ASTM D412 dogbone C tensile tests \cite{pinskier2022jammkle,Liow2023polyjet}. An incompressible hyperelastic Yeoh model was fitted to the Shore A 30 data, resulting in Yeoh parameters $C_1=\SI{1.2e-2}{\mega\pascal}$,  $C_2=\SI{-1.0e-4}{\mega\pascal}$, $C_3=\SI{6.2e-4}{\mega\pascal}$ \cite{pinskier2022jammkle}. Owing to its negligible nonlinearities, the Shore A 85 material was modelled as linearly elastic with modulus $E_{A85}=\SI{11.51}{\mega\pascal}$ \cite{Liow2023polyjet}. The material properties of of Shore D Vero material were obtained directly from the manufacturer's datasheet \cite{materialiseVeroPolyJet}. Viscoelasticity of the materials was negated for tractability. Contact between (I) stiff to stiff fibres (Shore A 85/Shore A 85), (II) soft to stiff fibres (Shore A 30/Shore A 85), and (III) soft to soft fibres (Shore A 30/Shore A 30) were modelled using the penalty contact method with Coulomb friction. The coefficients of friction for the three material parings are estimated as $\mu_{A30-A30}=0.74$, $\mu_{A30-A85}=0.75$, and $\mu_{A85-A85}=0.99$, respectively \cite{pinskier2022jammkle}. These friction values were estimated by fitting the parameters to experimental jamming data \cite{pinskier2022jammkle}. A non-linear least squares fit was performed using the trust-region-reflective algorithm \cite{pinskier2022jammkle,Branch1999}.

From the numerical simulation results in Fig. \ref{fig:fea} (a), the interactions between neighbouring fibres for both the tendon's unjammed and jammed states could be observed. While unjammed, deformation arises mainly from elongation of the softer Shore A 30 fibre sections. When jammed, interfibre friction prevents the stiffer Shore A 85 sections from sliding across each other and forces them to deform instead. Hence, the total deformation is distributed more uniformly along the length of the tendon.

The force-displacement curves obtained through numerical FEM simulation when compared real experimental results, were shown to accurately predict the stiffness and damping behaviour of the tendon, as shown in Fig. \ref{fig:fea} (b)(c). Unjammed, the tendon exhibits roughly a linear elastic behaviour with low stiffness. When the tendon is jammed, the stiffness increases along with damping capacity of the tendon, due to increase in interfibre friction and hence, shear force required for slip between fibres. The numerical model was able to capture the tendon's behaviour in frictional mode, in which the displacement under jamming occurs in three distinct phases: (I) pre-slip, whereby the fibres stretch together as a single cohesive unit, resulting in a high initial stiffness, followed by (II) the slip point, in which the shear force is large enough to exceed the friction limit and the fibres begin to slip, and (III) post-slip, in which fibres have slipped at all possible points along their interfaces. This leads to a reduction in tensile stiffness and a plastic deformation behaviour. Upon slip, energy is dissipated through friction between fibres and the entire structure behaves more like a damper. The numerical model was able to capture the stick-slip behaviour of the fibres, as observed by the 'jagged' region of the force-displacement plot. However, in reality, the fibre bundle does not compress evenly throughout and reorients slightly during extension, hence the slip point is not as clearly visible. This is especially seen for the smaller tendons and smoother curves are observed instead, as shown in Fig. \ref{fig:tendon} (c) (A-E). For larger tendons with larger fibre diameters and number of layers, such as Fig. \ref{fig:tendon} (c) (F-J), slip points are more obvious due to greater the contact area and hence, shear force required for slip. 

Using the simulation force-displacement data, a low-order polynomial regression fitting was subsequently performed as shown in Fig. \ref{fig:fea} (d). Fourth order polynomials were used to approximate the loading and unloading curves of the hysteresis loops for both unjammed and jammed states. These results will be used in later Section. \ref{section-multibodyfea} to simulate the characteristics of the tendon in a multibody dynamic model of the JEG.

\begin{figure}[t!]
\centerline{\includegraphics[width=0.9\columnwidth]{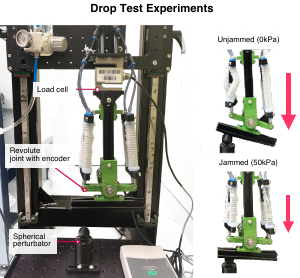}}
\caption{Left: Experimental setup to perform drop tests. Right: Behaviour of the joint after collision when unjammed vs jammed.}
\label{fig:droptestsetup}
\end{figure}

\section{Joint Impact Loading Experiment}
\label{section-damping}
One of the key advantages of introducing compliant elements in legged robots is to dampen the effects of sudden impacts, 
reducing joint load, and mitigating  abrupt movements that could affect stability of a system. The tunable mechanical properties of the fibre jammed tendons offer many advantages, which include the the varying stiffness and damping of a joint, while acting as a buffer or 'circuit breaker' to a mechanical system to prevent damage to the actuators. The first part of the experiments focuses on evaluating the performance of the proposed fibre jammed tendons by analysing the impact response behaviour of revolute joint coupled with a simple antagonistic lever-type mechanism, as shown in  Fig. \ref{fig:droptestsetup}.

\subsection{Drop Test Experimental Setup}
Two 150\,mm aluminium profile extrusions were used to mock-up the foot and lower half of a robotic leg. A tension-compression load cell (Laumas SA60, 60\,kg load capacity) was mounted to one end of the aluminium beam to measure the forces travelling up the leg upon impact. A rotary quadrature encoder (Yumo E6B2-CWZ3E-1024, 1024 pulses per rotation resolution) was used to measure the joint angular velocity. Impact force and angular velocity measurements were collected using a Data Acquisition System (National Instruments USB-6211) at a sampling rate of 1000\,Hz and results were smoothed using a Savitzky-Golay filter.

The mock-up leg was subsequently attached to a drop-test system, which includes a vertical sliding platform on linear rails. The platform mass and mock-up leg is approximately 2.0\,kg in total. A rigid spherical perturbator was mounted at the base of the system, to induce a perturbation or impact on the falling system. For all experiments, the mock-up leg was dropped from a height of 50\,mm above the perturbator (distance measured from the tip of the perturbator to the foot pad of the leg). A Rocker 500 vacuum pump with an adjustable pressure valve was used to set the jammed pressure of the tendons. Each drop test was repeated five times and the average force and angular velocity results were computed.

\begin{figure}[t!]
\centerline{\includegraphics[width=\columnwidth]{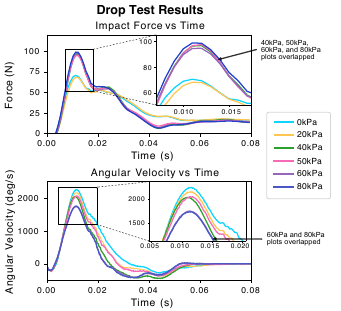}}
\caption{\textbf{Drop test results.} Top: Force-time plots depicting distinct changes in impact force over time, characterised by clear distinctions between low (0 and 20\,kPa) and high pressure (40, 50, 60, and 80\,kPa) settings. Bottom: Joint angular velocity-time plots showing that damped response is adjustable with pressure. }
\label{fig:droptestresults}
\end{figure}

\subsection{Drop Test Experimental Results}
Results from the drop test are shown in  Fig. \ref{fig:droptestresults}. For higher pressures beyond 40\,kPa, it was observed that the peak impact forces of up to 99\,N were observed and the plots showed similar responses. In contrast, at lower pressures such as 0\,kPa and 20\,kPA, impact forces were greatly attenuated, with peak impact forces effectively reduced by approximately 30\%. At low pressures, the tendons have relatively low stiffness and behave more as shock absorbers. However, the lower tendon stiffness and consequently joint stiffness resulted in a more uncontrolled drop, with higher joint angular velocities observed upon impact, as seen in the angular velocity-time plots. At higher pressures, the tendons were found to be more effective in slowing down the joint and preventing large unwanted motions due to perturbations acting on a system. This is because at higher pressures, the damping capacity of these tendons increases in conjunction with stiffness and greater energy is dissipated from the impact. Another interesting observation from the results worth noting is that while angular velocity of the joint can be modulated by vacuum pressure, the impact force response is approximately binary. This can be attributed to the fact that stiffness change of fibre jammed structures in response to pressure is mostly binary, while their damping properties could be further adjusted with pressure. 

The combination of inherent stiffness and damping of the tendons, which is tunable at material level by altering the geometric properties and configuration of the fibres, is advantageous over traditional variable stiffness systems, such as VSAs, in which only stiffness can be tuned. Furthermore, the tendons can be easily fabricated at a low cost with the design scalable for various applications. In the following sections, we demonstrate the versatility of the proposed fibre jammed tendons to allow adaptation and stable interactions between robots and their environment. We develop a novel design of a robotic leg, known as the JEG, with fibre jammed structures used in an antagonistic configuration to modulate stiffness and damping on the knee and ankle joints.

\begin{figure*}[!t]
\centerline{\includegraphics[width=2\columnwidth]{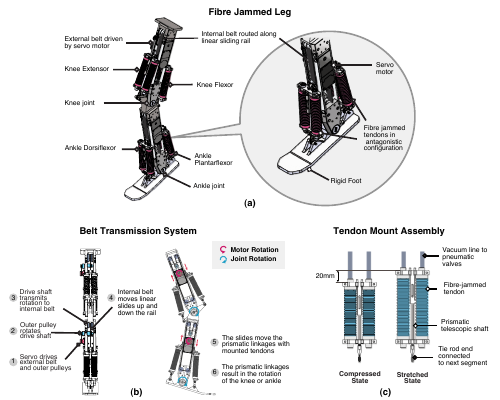}}
\caption{(a) Design and assembly of the fibre jammed leg. (b) Belt transmission system of the JEG. Motion from the servo motors transmitted by the belts to move the linear slides, which push and pull the prismatic linkages to rotate the knee and ankle joints. (b) The fibre jammed tendons are mounted in pairs. The prismatic joint limits the extension of the tendon in the longitudinal direction to a maximum of 20\,mm. A vacuum line is inserted for each tendon through an inlet bore located at the end caps of the tendons.}
\label{fig:design}
\end{figure*}

\section{Design of Fibre Jammed Leg}
\subsection{Mechanical Design of the JEG}

The JEG consists of three single Degree-Of-Freedom (DOF) rotational joints whose motion is constrained to the sagittal plane, as shown in Fig. \ref{fig:design} (a). The JEG has a hip joint for flexion-extension, knee joint for flexion-extension, and ankle joint for plantarflexion-dorsiflexion. The main body of the femur and tibia were constructed using aluminium frames, which act as linear rails, held together by 3D printed Acrylonitrile Styrene Acrylate (ASA) brackets. These brackets also act as revolute joints for the knee and ankle. The hip joint is actuated with a servo motor (ROBOTIS Dynamixel Pro H42-20-S300-R), while the knee and ankle joints are actuated using a timing belt transmission system with servo motors (ROBOTIS Dynamixel XM540-W270-T/R), which can produce torques of up to 12.9\,Nm.

Each servo drives an external belt connected to an outer-pulley, which is connected to the drive-shaft. The drive-shaft rotates the internal-pulley, which drives the internal belt. The design of the belt transmission system, along with routing path of the belt is shown in Fig. \ref{fig:design} (b). The internal belt drives the prismatic telescopic shafts to create a linear motion and hence, resulting in the compression or extension of the fibre jammed tendons. Each of the four prismatic joints were constructed out of a machined shaft, which slides along a rigid cylindrical sleeve. One end of the prismatic joint (the sleeve) was mounted to a linear slider, while the other end (the machined shaft with tie-rod end) was mounted to the neighbouring section of the JEG. A small aluminium plate was used to fix the internal belt onto the linear slide. Hence, movement of the belt results in equal displacement of the slides in the opposite direction along the length of the aluminium frame, which also acts as a guiding rail. When the servo motors are idle, the slides are fixed in place due to the frictional resistance and tension of the belt. The belts were selected through load capacity calculations by taking into account the working loads, which includes the spring forces from the tendons and impact forces encountered during the preliminary walking experiments. Performance of the belt was verified during these preliminary experiments in an iterative process, to ensure that the selected belts were able to handle the loads for this application. An adjustable belt tensioner was also used to ensure that the belt remains in tension to avoid slippage.

\subsection{Placement of the Fibre Jammed Tendons}

The limb movement of humans and most animals enabled by antagonistic pairs of muscles, which work in opposition to perform work. Inspired by this mechanism, an antagonistic lever type mechanism was adopted for the joints on JEG. The JEG uses four monoarticular tendon bundles, with a pair of tendons placed in an antagonistic arrangement on each side of the knee and ankle joints. The tendons are placed in this position to simulate major muscle-tendon groups, such as the hamstrings, quadriceps, shin and calve muscles. These pairs of tendons can be categorised as:
\begin{enumerate}
    \item \textbf{Knee Extensors} or the quadriceps muscles, which include the rectus femoris and vasti. Their role include decelerating flexion of the knee as a result of the ground reaction forces during the heel-strike or loading response phase. The knee extensors also function to stabilise the knee after the push-off or pre-swing phase, due to the residual activity of the ankle plantarflexors.
    \item \textbf{Knee Flexors}, which are the hamstring muscles and antagonist muscles of the quadriceps. The muscles include the semimembranosus, semitendinosus, and biceps femoris. These muscles perform an important role in decelerating the extension of the knee and forward motion of the tibia after the midstance.
    \item \textbf{Ankle Dorsiflexors}, which include shin muscles such as the tibialis anterior. The role of the dorsiflexors is to provide foot clearance during the swing face and to control the plantarflexion of the foot during heel-strike, as a result of transfer of the body weight.
    \item \textbf{Ankle Plantarflexors} are the calf muscles. The gastrocnemius and soleus are the major muscles in this group and are attached to the Achilles tendon. They function as a shock absorber during running and walking, as well as to generate thrust during the push-off phase.  
\end{enumerate}

 The pairs of tendons on one side of the joint work antagonistically with the opposite pair, and collectively define the joint stiffness. The knee extensor and knee flexor work antagonistically to impose a stiffness on the knee joint, while ankle dorsiflexor and ankle plantarflexor together impose a stiffness to the ankle joint. Each joint, and hence, each tendon, affects the mechanical impedance on the movement of the leg. Stretching in one tendon will result in compression of the opposite tendon in response to an external force, such as the ground reaction or when the leg collides with an obstacle. The tendons, along with the prismatic joints behave as buffers or passive decoupling mechanisms between the belt transmission and neighbouring sections, such as the femur from the tibia and the foot from the tibia. The tendons provide mechanical compliance and damping, due to their hyperelastic material properties, allowing the foot to passively adapt to an obstacle during collision without resulting in damage to the timing belt or motors. The tendons can then be subsequently stiffened through jamming to improve traction of the leg and generate thrust. When the obstacle or load is removed, the tendons will relax and guide the leg back to its original trajectory. On the other hand, when the leg is air-borne, the prismatic joints are free to move with the linear slides with minimal resistance from the tendons. This is because the tendons are only stretched or compressed when the leg is in contact with the ground or when loaded with an external force (e.g during the stance phase of locomotion). 
 
 The pairs of tendons were mounted to the prismatic joint as shown in Fig. \ref{fig:design} (c), with end caps clamped the sleeve and shaft of the joint. The prismatic joints were also designed with mechanical limits to ensure that the maximum extension is 20\,mm, to ensure that the tendons were within operational loading conditions. A Rocker 500 vacuum pump and pressure regulator were used to jam the tendons. Each tendon mechanism had a separate vacuum line running from the master point to the mounting point on the tendon holder. All tendons were jammed to 50\,kPa, as there was little force increase with an increase beyond that pressure. It is important to note that the Rocker 500 vacuum pump was utilised due to its availability in the lab. Nevertheless, there are alternative lightweight options such as micro DC vacuum pumps that are capable of generating comparable pressures.

\begin{table}[!t]
\centering
\caption{\label{tab:parameter} Simulation Variables and Parameters}
\def\arraystretch{1.75}
\begin{tabular}{l l c l}
\toprule
\textbf{} & \textbf{Description} & \textbf{Value} & \textbf{Unit} \\
\midrule
$\theta_{hip}$ &  Touchdown Hip Angle & 14 & [$^\circ$]\\
$\theta_{knee}$ &  Touchdown Knee Angle & 17.5 & [$^\circ$]\\
$\theta_{ankle}$ &  Touchdown Ankle Angle & 80 & [$^\circ$]\\
$g$ &  Gravity & 9.8066 & [$m/s^2$]\\
$\mu_{F}$ & \makecell[l]{Foot-ground Friction\\ Coefficient}& 0.4 & \\
$x_{1}$ & \makecell[l]{Knee Extensor Prismatic\\  Joint Displacement} & -10 to 10 & [mm]\\
$x_{2}$ & \makecell[l]{Knee Flexor \\Prismatic Joint\\ Displacement} & -10 to 10 & [mm]\\ 
$x_{3}$ & \makecell[l]{Ankle Dorsiflexor \\Prismatic Joint\\ Displacement} & -10 to 10 & [mm]\\ 
$x_{4}$ & \makecell[l]{Ankle Plantarflexor \\Prismatic Joint\\ Displacement} & -10 to 10 & [mm]\\ 

\bottomrule

\end{tabular}

\end{table}

\begin{figure*}[!t]
\centerline{\includegraphics[width=2\columnwidth]{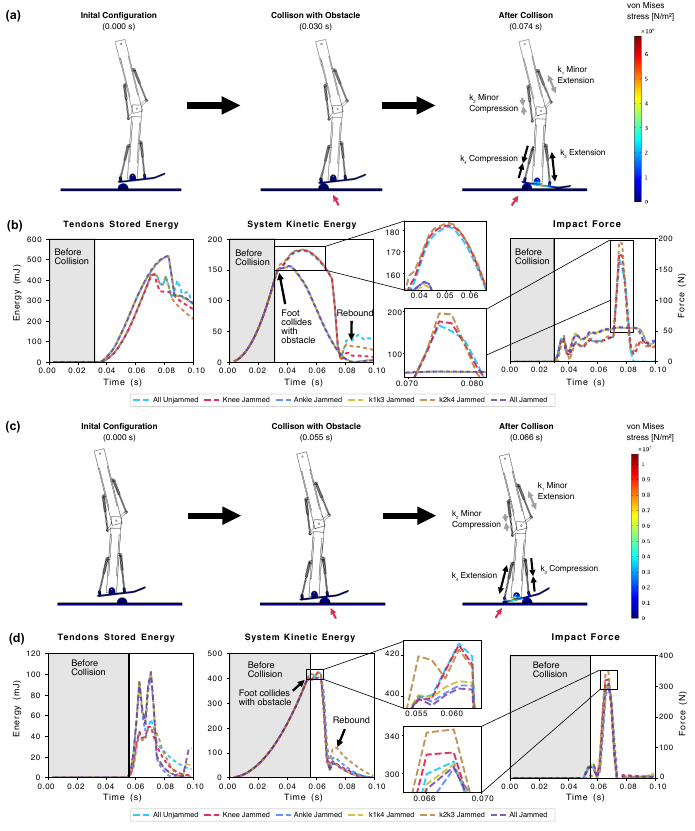}}
\caption{\textbf{Simulation of toe-down collision: } (a) Initial touchdown configuration of the leg and resulting dynamic response after collision with obstacle at heel, subsequent lowering of the toes to ground plane. (b) Total stored energy in the tendons, total system kinetic energy, and impact force when (1) all tendons unjammed, (2) only knee tendons jammed, (3) only ankle tendons jammed, (d) only extending tendons ($k_{1}$$k_{3}$) jammed, (5) only compressing tendons ($k_{2}$$k_{4}$) jammed, and (6) all tendons jammed. \textbf{Simulation of toe-up collision: }(c) Initial touchdown configuration of the leg and resulting dynamic response after collision with obstacle at forefront of foot, subsequent lowering of the heel to ground plane. (d) Total stored energy in the tendons, total system kinetic energy, and impact force when (1) all tendons unjammed, (2) only knee tendons jammed, (3) only ankle tendons jammed, (4) only extending tendons ($k_{1}$$k_{4}$) jammed, (5) only compressing tendons ($k_{2}$$k_{3}$) jammed, and (6) all tendons jammed.}
\label{fig: energy-sim}
\end{figure*}

\section{Numerical Model and Design of the Fibre Jammed Leg}
\label{section-multibodyfea}
From the joint impact loading experiment in Section. \ref{section-damping}, we observed the remarkable adaptability of the fibre jammed tendons in imparting both variable stiffness and damping properties to a revolute joint. Humans are able to modulate the gait and stiffness of their legs to efficiently adapt to different terrains. This is done through the synergistic contractions of muscles. However, there is great uncertainty in how this exactly translates in a robotic system. Hence, to gain further insight on how the fibre jammed tendons could be exploited in mulit-jointed robotic leg as a whole, we propose a simplified multibody dynamic FEM model of the JEG, which will allow us to analyse the change in energy of the system prior and after a collision with an unexpected obstacle. By using the mechanical properties of the tendons previously computed in Section \ref{section-tendonfea}, we vary the jammed states of different pairs of tendons to evaluate the performance of the leg as a whole in reducing impact forces and energy for toe-down and toe-up rotational perturbations.

\subsection{FEA Simulation Methods}
A simplified model of the JEG was designed in CAD and imported into COMSOL Multiphysics to perform a time-dependent multibody dynamics 3D FEM simulation, with variables and parameters shown in Table. \ref{tab:parameter}. The FEM model's materials were set based on the JEG prototype, with the femur and tibia modelled as aluminium, the prismatic joints as steel, the plastic foot as ASA, and the ground with the obstacle object as aluminium. To reduce computational expense, only the foot and the ground were modelled as linear elastic, while the rest of the domains were considered rigid with inertial terms considered. The hip, knee, and ankle joints were modelled as 1-DOF revolute joints. The belt transmission system was simplified to only include the linear slides, which are locked in position to simulate the non-backdrivability of the belt. The tendons and their telescopic shafts were modelled as prismatic joints with spring-damper subnodes
between their shaft and the sleeve domains to apply elastic and viscous forces to control the movement of the joint. The spring-damper's mechanical properties were expressed as non-linear force-extension equations, to simulate the hyperelastic behaviour of the tendons, in both their jammed and unjammed states. This was done by utilising the previously simulated force-displacement data and performing a low-order
polynomial regression fitting, as described in Section. \ref{section-tendonfea}. As the tendons only provide a tensile force during extension, stiffness and damping were only implemented when there was a positive displacement (extension) of the prismatic joints. When the prismatic joints undergo a negative displacement (compression), the fibres will move and rearrange themselves, leading to buckling of the tendon.
From the tendon characterisation experiments, the compressive forces produced by the tendons were observed to be significantly lower than their tensile forces (less than 5\% for the largest tendon sample, the 3.0mm diameter Hex 4 Layer tendon). Therefore, for simplification, the stiffness and damping of the tendon during compression were assumed to be negligible. For ease of reference, the knee extensor was denoted as $k_{1}$, the knee flexor as $k_{2}$, the ankle dorsiflexor as $k_{3}$, and ankle plantarflexor as $k_{4}$.

The obstacle object was modelled as a hemisphere of 40\,mm diameter, to create a spherical contact upon collision with the foot. Contact pairs were defined between the bottom surface of the foot and the top surfaces of both the obstacle and ground, with a friction coefficient of $\mu_{F}= $ 0.4 using the Coulomb friction model. The dynamic penalty method was selected due to its stability and robustness for complex dynamic impact problems. The collision with obstacle object will induce a rotational perturbation on the leg, resulting in the stretch of different tendons depending on the direction (clockwise or anti-clockwise) of joint rotation. As the JEG is symmetrical about the sagittal plane, we focused mainly on perturbations which resulted in movement of the leg in the anterior-posterior direction. In the simulation, two studies were performed, a (1) toe-down collision, where the obstacle object was located before the ankle and impact would occur around the heel, and a (2) toe-up collision, where contact with the obstacle would occur in front of the foot.  As the entire geometry of the leg is symmetrical about the sagittal plane, the centre axis of the hemispherical obstacle was only displaced in the horizontal direction, along the plane. The JEG was dropped at a height of 5\,mm measured to the top surface of the obstacle to the base of the foot. The velocity of the foot at the point of collision is approximately 0.3\,m/s for toe-down simulations and  0.6\,m/s for toe-up simulations, with values slightly differing between different jammed configurations. A parametric sweep for each of the two studies were performed, with the state of the spring-damper subnode of the prismatic joints set to either  unjammed or jammed. The change in stored energy in the tendons, kinetic energy of the whole system, and impact force upon collision were measured in the simulation before and after impact with the obstacle object. It is important to note that the simulation is modelled as an inelastic collision and hence, energy is kinetic energy is not conserved. Therefore, the computed kinetic energy results accounts for energy loss due to material deformation and friction due to mechanical contact, in addition to energy dissipated by the tendons. 

The mesh of the 3D FEA model for the toe-down and toe-up dynamic study consist of 71505 elements and 72027 elements respectively. To improve accuracy of the computed forces, mesh refinement was carried out in the regions where contact occurs, namely the underside of the foot and top of the obstacle object. Each parametric sweep took approximately 120 minutes and computations were carried out using a 3.10GHz Intel Core i9-9960X X-series processor with 64GB RAM.

\subsection{FEA Simulation Results}

\textbf{Toe-down Collision Results:}
Upon collision with an object at the heel, the reaction force resulted in an immediate ankle plantarflexion or toe-down motion of the foot, accompanied by a knee flexion, as seen in Fig. \ref{fig: energy-sim} (a). This results in an initial extension of the ankle dorsiflexor ($k_{3}$), followed by a minor extension of the knee extensor ($k_{1}$). From the energy plots in Fig. \ref{fig: energy-sim} (b), energy stored in the tendons were the greatest when: only the ankle tendons were jammed  ($k_{3}$$k_{4}\,Jammed$), only the extending tendons jammed ($k_{1}$$k_{3}\,Jammed$), or when all tendons were jammed (\textit{All Jammed}), with all three configurations showing similar energy profiles. For these three configurations, significantly lower impact kinetic energies were observed at touchdown, followed by the system kinetic energy settling to zero after the 0.085\,s mark. A higher damping property of the leg leads to reduced impact kinetic energy upon initial collision with the obstacle. Furthermore, higher damping is also beneficial in preventing rebound and unwanted oscillations at touchdown. The greatest rebound was seen with the \textit{All Unjammed} configuration. This behaviour can also be attributed to the low stiffness and damping of the joints, which resulted in the uncontrolled movement of the joints in response to the impact. Similarly, large rebound energies within the system were also seen when only compressing tendons jammed ($k_{2}$$k_{4}\,Jammed$) and when only knee tendons were jammed ($k_{1}$$k_{2}\,Jammed$). For these three configurations, extremely high impact forces were seen after the collision (at 0.074\,s mark), when the forefront of the foot touches the ground. The peak impact force \textit{All Unjammed}, $k_{2}$$k_{4}\,Jammed$, and $k_{1}$$k_{2}\,Jammed$ configurations are 166\,N, 176\,N, and 195\,N respectively. In contrast, the $k_{1}$$k_{3}\,Jammed$, $k_{3}$$k_{4}\,Jammed$, and \textit{All Jammed} configurations showed smoother force-time plots no distinguishable peak forces. This indicates better impact attenuation and adaptivity of the leg, which is beneficial to mitigate rebound from the impact.

\textbf{Toe-up Collision Results:} In our previous toe-down collision study, it was observed that the knee extensor ($k_{1}$) and ankle dorsiflexor ($k_{3}$) were the main tendons in extension. However, when the nature of the perturbation or location of the perturbator was changed, the limb and joint dynamics were found to change as well in simulation. When the leg collides with an object in front of the foot, an ankle dorsiflexion or toe-up motion occurs, followed by a knee flexion, as shown in Fig. \ref{fig: energy-sim} (c). Therefore, this results in the extension of the ankle plantarflexor ($k_{4}$), followed by the minor extension of the knee extensor ($k_{1}$). Similarly to the previous study, when these tendons in extension were jammed, the ($k_{1}k_{4}$), only ankle tendons jammed ($k_{3}k_{4}$), or when all tendons are jammed (\textit{All Jammed}), the energy stored in the tendons surpassed all other configurations, as presented in Fig. \ref{fig: energy-sim} (d). For these configurations, lower kinetic energies were seen at impact, followed by lower rebound energies. The \textit{$k_{2}$$k_{3}$ Jammed} configuration showed the greatest increase in kinetic energy after rebound, and this is followed by only ankle tendons jammed ($k_{3}$$k_{4}$) configuration. For these two configuration, high impact forces were also observed of 346\,N and 322\,N respectively. In contrast to the toe-down collision study, energy and impact force results were less disparate between configurations. This occurs because in the toe-up collisions, the initial touchdown pose of the leg and position of the obstacle object lead to less displacement of the tendons compared to toe-down collisions.

In summary, our results show that the jamming of different tendons could affect the dynamics of a system, which could be beneficial for disturbance rejection and consequently, improving locomotion stability in complex environments. We showed that \textit{All Jammed} configuration performed much more favourably than \textit{All Unjammed} configuration in lowering the kinetic energy of the impact and forces upon collision. Jamming of extending tendons provided a similar response to the \textit{All Jammed} configuration,  thereby necessitating reduced air consumption for the system and enabling greater energy efficiency. Another interesting observation from the results revealed that jamming only the ankle tendons proved significantly more effective than jamming only the knee tendons, which warrants further investigation on the roles and contribution of each tendon towards the compliance behaviour of the leg.

\section{JEG Rotational Perturbation Experiment}

Understanding  the dynamic behaviour of a multi-joint system, such as a robotic leg, compared to a single joint is non-trivial, as seen from the multibody dynamic simulations in Section. \ref{section-multibodyfea}. The relative importance of each of the four pairs of tendons in negotiating perturbations remains unsolved. How important are the stiffness of the knee tendons compared to the ankle tendons? How important are the extensors over the flexors? In this experiment, we perform a quantitative analysis to determine which tendon plays the most critical role in contributing to the overall stiffness of the JEG. Rotational perturbations are one of the most common types of perturbations seen in daily life, which causes displacement of the knee and ankle joints and consequently result in instability of a system.  We perform an indentation test against the foot of the JEG to measure its resistance against an induced perturbation for different jammed combination of tendons. As force is generally correlated with the stiffness of a structure, force sensor readings from the indenter were collected for all jammed combinations and pairwise cross-comparisons were conducted to determine the efficacy of jamming a targeted tendon. A higher indentation force will indicate a greater overall leg stiffness, while a lower indentation force will indicate a lower leg stiffness or greater compliance and therefore, adaptation to an external applied load.

\begin{figure}[!t]
\centerline{\includegraphics[width=\columnwidth]{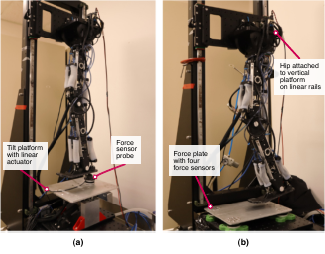}}
\caption{(a) Perturbation experimental setup with an actuated tilting support platform. (b) Walking experimental setup with force plate. }
\label{fig:experiment}
\end{figure}

\subsection{Perturbation Experimental Setup}

A tilting support surface actuated by a linear actuator with potentiometer feedback (Actuonix Micro Linear Actuator P16-50-64-12-P) was used to create the pitch plane perturbations, as shown in Fig. \ref{fig:experiment} (a). The tilting platform utilises a 'see-saw' mechanism, in which a rigid aluminium plate rotates about a bearing supported shaft located at the centre. A 3-axis force sensor (OptoForce OMD-45-FH-2000N) with a dome of diameter 30\,mm and height 20\,mm was mounted on the plate to simulate a point load upon contact with the foot. The dome will ensure that forces are transmitted through the centre of the force sensor. Measurements obtained from the force sensor will indicate the 'opposing force' produced by the JEG against the loading dome object, which acts as the perturbator. It was assumed that for the same rotational displacement, the opposing force the leg exerts towards the perturbator is proportional to the overall leg stiffness. To minimise slip and consequently, the loss of force information due to displacement of the foot, a 3D printed sock of Shore A 60 was used.

The JEG was initially configured to a touchdown or heel-strike pose. The servo motors were left powered on but are unactuated to ensure there was no back-driving of the belt transmission upon loading of the leg. Additionally, it was essential for the motors to remain unactuated to effectively decouple the tension forces generated by the tendons from the forces exerted by the actuators, as it will affect results of the experiment. The same types of tendons were used (2.0 mm fibre diameter 4 Hex layer) for the knee extensors, knee flexors, ankle dorsiflexors, and ankle plantarflexors. These specific tendons were selected based on the loading capacity of the belt drive actuation system. The four pairs of tendons were jammed in different configurations, to achieve varying levels of stiffness. As each tendon pair has two states, either (1) unjammed or (2) jammed at 50 kPA, there are hence, $2^{4}$ possible combinations in the permutation space. For each of the 16 possible permutations, two rotational perturbation experiments were performed and repeated five times, resulting in a total of 160 loading cycles. Between each loading cycle, the vacuum is switched off and atmospheric pressure is allowed to return into the membranes. A waiting period of 5 minutes was allowed to ensure that the tendons return to their original configurations, to avoid bias in the data. The setup of the rotational perturbation experiments are as follows:
\begin{figure}[!t]
\centerline{\includegraphics[width=0.8\columnwidth]{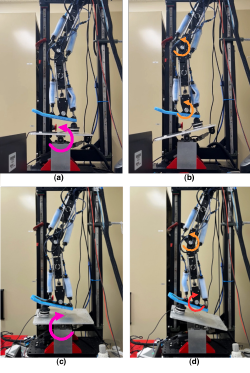}}
\caption{\textbf{Toe-down perturbation experiment: } (a) The force sensor was raised by the tilting platform and contact is made with the heel. (b) Loading on the heel results in a major ankle plantarflexion and flexion of the knee, as shown by the anti-clockwise arrows. \textbf{Toe-up perturbation experiment: }(c) The force sensor was raised by the tilting platform against the metatarsal area of the foot. (d) Loading on the front of the foot results in an ankle dorsiflexion and slight flexion of the knee, in which the ankle rotates clockwise and knee rotates anti-clockwise.}
\label{fig:perturbexperiment}
\end{figure}

\begin{enumerate}
    \item \textbf{Toe-down Perturbation: } The force probe was mounted 60\,mm distance from centre of plate before the ankle. The tilting surface rotates in the anti-clockwise direction up to 15$^{\circ}$, resulting in an increasing load applied to the heel, as shown in Fig. \ref{fig:perturbexperiment} (a)(b). 
    
    \item \textbf{Toe-up Perturbation:} The force probe was mounted 80\,mm distance from centre of plate after the ankle. The tilting surface rotates in the clockwise direction up to 15$^{\circ}$, resulting in an increasing load applied to the front of the foot, as shown in Fig. \ref{fig:perturbexperiment} (c)(d). 
\end{enumerate}

The highest recorded resultant force during leg loading was obtained, and the mean was calculated from the five performed cycles, with results shown in
Table. \ref{tab:perturbation}. In both the toe-down and toe-up experiments, the mean force computed was the lowest when all tendons were unjammed, while it was the highest when all tendons were jammed, as compared to other configurations. However, when considering configurations with only two tendons jammed, it was observed that certain recorded forces exceeded those in which three tendons were jammed. Hence, subsequent analysis of the contribution of each tendon pair towards opposing either a toe-down or toe-up perturbation is required. Hence, individual contributions of the tendons therefore quantified by computing the difference in measured forces between the configuration in which the targeted tendon pair (e.g $k_{1}$) was jammed with the configuration in which the targeted tendon pair was unjammed (e.g ${k_{1jam}/k_{2jam}/k_{3unjam}/k_{4unjam}}$ vs ${k_{1unjam}/k_{2jam}/k_{3unjam}/k_{4unjam}}$). This analysis was performed for each of the targeted tendons and the mean paired difference was computed by summing the individual difference from the eight pairs, and then dividing by eight.

\begin{table}[!t]
\centering
\caption{\label{tab:perturbation} Perturbation Test Results. }

\begin{tabular}{ |p{0.2cm}|p{0.2cm}|p{0.2cm}|p{0.2cm}|p{2.2cm}|p{0.05cm}|p{1.8cm}|}

\hline
\textbf{$k_{1}$} & \textbf{$k_{2}$} & \textbf{$k_{3}$} & \textbf{$k_{4}$} & \textbf{Toe-down Mean Force (SD) [N]}  &\cellcolor{black!25}& \textbf{Toe-up Mean Force (SD) [N]}\\
\hline
&&&& \cellcolor{blue!5} 48.7 (0.597) &\cellcolor{black!25} & \cellcolor{blue!5} 11.4 (0.394)\\
\hline

\cellcolor{black!}&&&& \cellcolor{blue!15} 50.0 (1.496) & \cellcolor{black!25} & \cellcolor{blue!15} 19.1 (1.371)\\
\hline
&\cellcolor{black!}&&& \cellcolor{blue!15} 48.8 (0.470) & \cellcolor{black!25} & \cellcolor{blue!15} 16.8 (0.433)\\
\hline
&&\cellcolor{black!}&& \cellcolor{blue!15} \textbf{54.1 (0.404)} & \cellcolor{black!25} & \cellcolor{blue!15} 
 21.2 (1.264)\\
\hline
&&&\cellcolor{black!}& \cellcolor{blue!15}  50.1 (0.472) & \cellcolor{black!25} & \cellcolor{blue!15} 
 \textbf{29.1 (0.440)}\\
\hline
\cellcolor{black!} & \cellcolor{black!}&&& \cellcolor{blue!25} 60.4 (0.210) &\cellcolor{black!25} & \cellcolor{blue!25} 18.8 (0.997)\\
\hline
\cellcolor{black!} && \cellcolor{black!} && \cellcolor{blue!25} \textbf{74.8 (0.759)} &\cellcolor{black!25} & \cellcolor{blue!25} 16.2 (0.950)\\
\hline
\cellcolor{black!} &&& \cellcolor{black!} & \cellcolor{blue!25} 62.0 (1.234) &\cellcolor{black!25} & \cellcolor{blue!25} 29.9 (0.840)\\
\hline
&\cellcolor{black!}&\cellcolor{black!} && \cellcolor{blue!25} 66.6 (2.340) &\cellcolor{black!25} & \cellcolor{blue!25} 22.5 (1.551)\\
\hline
&\cellcolor{black!}&&\cellcolor{black!}& \cellcolor{blue!25} 58.0 (2.349)
&\cellcolor{black!25} & \cellcolor{blue!25} 29.2 (0.527)\\
\hline

&&\cellcolor{black!}&\cellcolor{black!} & \cellcolor{blue!25} 62.0 (1.055)
&\cellcolor{black!25} & \cellcolor{blue!25} \textbf{32.5 (0.437)}\\
\hline

\cellcolor{black!}&\cellcolor{black!}&\cellcolor{black!}&& \cellcolor{blue!35} \textbf{69.7 (1.102)} &\cellcolor{black!25} & \cellcolor{blue!35} 25.0 (0.422)\\
\hline

&\cellcolor{black!}&\cellcolor{black!}&\cellcolor{black!}& \cellcolor{blue!35} 62.7 (0.708) & \cellcolor{black!25} & \cellcolor{blue!35} 32.4 (0.532)\\
\hline

\cellcolor{black!}&&\cellcolor{black!}&\cellcolor{black!}& \cellcolor{blue!35} 66.3 (0.344) &\cellcolor{black!25}& \cellcolor{blue!35} \textbf{32.6 (0.512)}\\
\hline

\cellcolor{black!}&\cellcolor{black!}&&\cellcolor{black!}& \cellcolor{blue!35} 58.9 (2.3076) & \cellcolor{black!25} & \cellcolor{blue!35} 30.2 (0.584)\\
\hline

\cellcolor{black!}&\cellcolor{black!}&\cellcolor{black!}&\cellcolor{black!} & \cellcolor{blue!50} 85.6 (2.270) & \cellcolor{black!25} & \cellcolor{blue!50} 36.4 (0.784)\\
\hline

\end{tabular}

\vspace{0.2cm}

\begin{minipage}{8cm}

\small *Values highlighted in bold indicate the configuration with the highest mean force for the same number of tendons jammed.

\end{minipage}

\vspace{0.2cm}

\begin{tabular}{|p{0.1cm}|p{1.2cm}|p{0.1cm}|p{1.2cm}|}
\hline
&Unjammed&\cellcolor{black!}& Jammed\\
\hline
\end{tabular} 

\end{table}

\subsection{Perturbation Experimental Results}
\begin{figure*}[!t]
\centerline{\includegraphics[width=2.05\columnwidth]{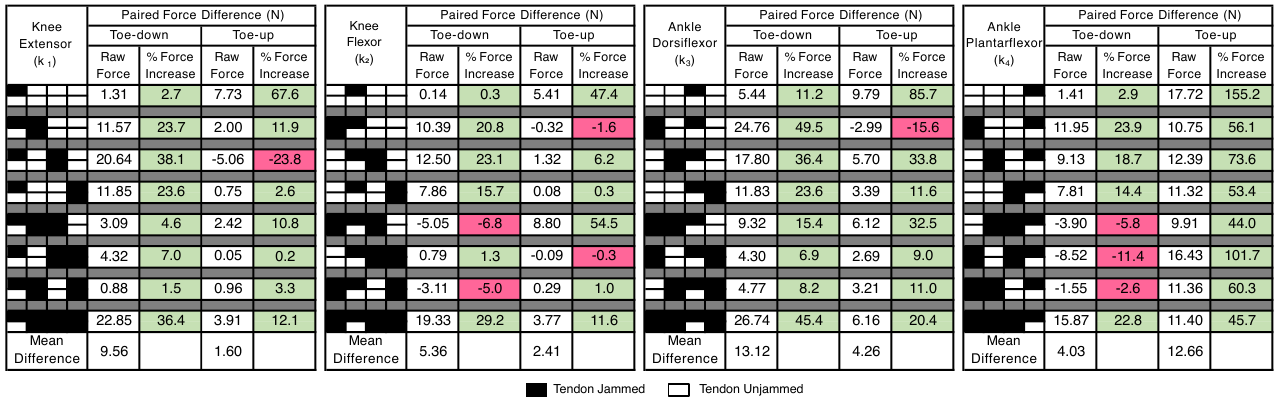}}
\caption{Resulting permutation space. The force difference between the configuration where the target tendon was jammed with the configuration where it was unjammed was calculated to quantify the force contribution of jamming the target tendon, while the rest of the tendons were in the same state. For each target tendon, the mean paired force difference indicates the overall contribution towards opposing either a toe-down or toe-up perturbation. Green values indicate a positive percentage (\%) force contribution, whereas red values indicate a negative (\%) force contribution upon jamming of that target tendon.}
\label{fig:averagediff}
\end{figure*}

\begin{figure}[!t]
\centerline{\includegraphics[width=\columnwidth]{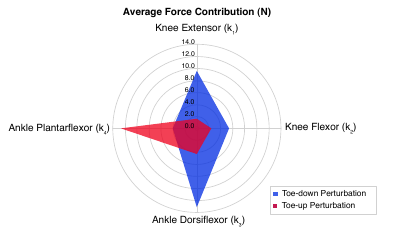}}
\caption{Mean force contribution of the tendons for (a) toe-down perturbation and (b) toe-up perturbation.}
\label{fig:contribution}
\end{figure}

With reference to Fig. \ref{fig:averagediff}, a positive force contribution is highlighted in green, indicating that jamming the targeted tendon resulted in a higher opposing force, and consequently, a greater overall leg stiffness towards the specific directional perturbation. In contrast, a negative contribution is highlighted in red, which indicates there is a loss in measured force or greater leg compliance when the targeted tendon was jammed. In this case, leaving the targeted tendon unjammed will be more favourable in opposing that directional perturbation. The mean force contribution from each of the tendons is plotted in Fig. \ref{fig:contribution}. Results for the toe-down perturbation experiment shows that the ankle dorsiflexor produced the highest contribution, a significant 13.1\,N. This is followed by the knee extensor, the knee flexor, and ankle plantarflexor producing a force contributions of 9.56\,N, 5.36\,N, and 4.03\,N respectively. These results can be explained by observing the dynamic response of the leg during the experiment in Fig. \ref{fig:perturbexperiment} (a)(b). Loading on the heel causes the foot to flatten and an ankle plantarflexion to occur. This is followed by flexion of the knee. Hence, the resulting joint rotation of the knee and ankle are in the same direction (anti-clockwise with reference to Fig.\ref{fig:perturbexperiment}). This causes the the knee extensors and ankle dorsiflexors to stretch, while the knee flexors and ankle plantarflexors compress. By jamming these two tendons, it was observed that the opposing force produced was significantly greater than any other configuration (with exception for the \textit{All Jammed} configuration), as shown in Table. \ref{tab:perturbation}. As the tendons produce a tensile force only with extension, and an insignificant compressive force due to buckling, it can be inferred that the stiffness of the joint (knee or ankle) is determined wholly by the tendons which are in tension. The toe-up perturbation experiment, however, shows opposite results, as the ankle plantarflexor provided the greatest opposing force contribution when loaded with the plate, 12.6\,N, followed by the ankle dorsiflexor, 4.26\,N. The knee tendons do not show as much contribution in comparison, with the knee extensor and knee flexor contributing forces of 1.60\,N and 2.41\,N, respectively. Through these findings, we gain a deeper understanding on the relative importance of each tendon to the stiffness of the leg as a whole. This deepens our understanding of how these jammed structures could be leveraged to alter limb compliance and improve performance of the leg as a whole.

\section{JEG Walking Experiment}
To demonstrate the performance of the fibre jammed tendons in a locomotion task, we perform walking experiments with the JEG with a force plate. The ground reaction forces measured from the force plate were used as a qualitative measure to evaluate the walking stability of the leg, along with its ability to dampen impacts at foot touchdown. Ground reaction forces (GRF) is a common method to quantify and analyse quality of a gait in human locomotion as well as in robotics, and will be used as a measurement for this experiment.

\subsection{Walking Experiment Setup}

The same test rig as with the previous experiments was used, but with the sliding platform unlocked, as shown in Fig. \ref{fig:perturbexperiment} (b). The hip joint servo motor was mounted on the vertical sliding platform to allow changing of the hip height. The platform was fabricated out of rigid plastic with mounting holes and was attached onto the pair of linear motion sliders. The vertical sliding platform itself does not have a motor. This is to ensure no additional forces contribute to the measured GRF, and that the platform is mainly used to facilitate the changing of the hip height. As the experimental setup comprises of only a single leg, clamps were used as mechanical limits on the linear rail, under the vertical platform. This is to ensure that the JEG lifts-off the plate during the swing phase, as there is no opposing leg to support its weight. However, during the stance phase, the vertical platform was free to lift from the thrust generated by the JEG. The mass of the JEG along with the vertical plate is 3.65\,kg in total.

A force platform was developed to measure the vertical and horizontal GRF generated by the JEG during the walking cycle. The force platform is 260 $\times$ 210\,mm in length,  which consists of a rigid flat aluminium plate with four 3-axis force sensors (OptoForce OMD-45-FH-2000N) mounted under each corner of the plate. The GRF produced by the leg is equal to the sum of the force readings obtained by all four sensors together. The forces produced by the JEG were sampled at 1000\,Hz using a DAQ (Optoforce) and were passed through a low-pass Gaussian filter to remove noise. The resultant GRF, $F_R$ was calculated from the $\sqrt{{F_z}^2 + {F_x}^2}$, where ${F_z}$ is the vertical and ${F_x}$ is the horizontal GRF obtained from the force plate.

\begin{figure}[!t]
\centerline{\includegraphics[width=0.98\columnwidth]{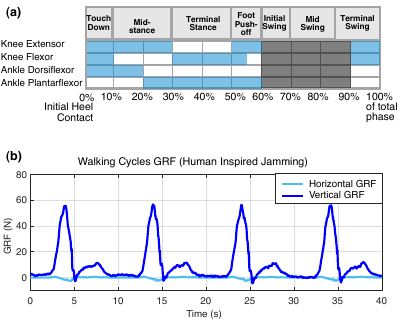}}
\caption{(a) Timing of applied jamming for knee extensors, knee flexors, ankle dorsiflexors, and ankle plantarflexors shown in percentage (\%) of the predefined gait cycle. In Case III, tendons were jammed with reference to human muscle activation data\cite{torricelli2016human,winter1991biomechanics}. (b) GRF-time data of the walking cycles when tendons were jammed with a human inspired muscle activation pattern.}
\label{fig:human}
\end{figure}

The human walking gait cycle consists of two phases: the stance phase, where the foot is in contact with the ground and the swing phase, where the foot is suspended mid-air. The stance phase can be further divided into the touchdown phase, in which the heel strikes the ground, the midstance, in which the foot flattens itself against the ground, and the push-off in which the foot pushes off and accelerates the tibia forward. As the tendons are only engaged when the foot is in contact with the ground, only the stance phase was focused on during the walking experiment. When the leg is airborne and no external forces is applied to the foot, there is little resistance from the tendons acting on the knee and ankle joints. Hence, the torque from the motors is directly utilised to actuate the leg without stretching or compression of the tendons. 

\begin{figure}[!t]
\centerline{\includegraphics[width=\columnwidth]{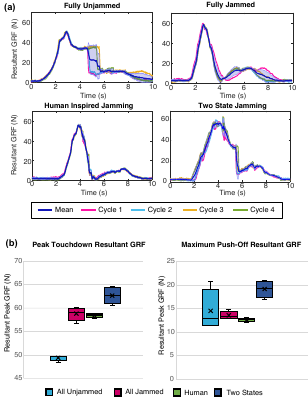}}
\caption{(a) Mean GRF obtained during stance phase when: (I) all tendons are unjammed, (II) all tendons jammed, (III) tendons activated based on human muscle activation synergy, and (IV) all tendons unjammed at heel-strike and jammed after midstance. (b) Left: Peak resultant GRF measured after touchdown of the foot. Right: Estimated peak push-off force for different jamming cases.}
\label{fig:walkingresults}
\end{figure}

A Bezier curve trajectory was prescribed on the foot and corresponding joint angles was calculated through inverse kinematics. The required servo motor angles were subsequently obtained by calculating the travel of the belt, for positional control with PID using the Dynamixel SDK in Python.  A heel-strike and push-off was implemented in the gait to ensure that the joint angular motion directions change throughout the gait, to ensure that the antagonistic pair of tendons are utilised. From the prescribed leg trajectory, the time at which the heel-strike, midstance and push-off has occurred was estimated and expressed in percentage gait cycle (\%). A PLC controller (CLICK PLC C0-11DD2E-D)  controlled the switching of the pneumatic solenoid valves connected to the vacuum to jam and unjam the four tendon pairs. With the current experimental setup, it takes approximately 300\,ms to reach final pressure when all valves are open (all tendons jammed) at the same time. The timing of activation the pneumatic valves were synchronised to relevant parts of the predefined gait cycle. The entire gait cycle duration was set to 10\,s and the same gait was used for all experiments. The walking test was performed using 2.0\,mm fibre diameter Hex 4 Layer tendons for four different jamming cases:
\begin{enumerate}
    \item \textbf{Case I: } All tendons unjammed throughout entire gait.
    \item \textbf{Case II: } All tendons jammed throughout entire gait.
    \item \textbf{Case III: } Tendons jammed with a human-like muscle activation pattern, as shown in Fig. \ref{fig:human} (a). The jamming sequence the tendons was estimated from previous human muscle activation pattern studies \cite{torricelli2016human,silder2013men,winter1991biomechanics}.
    \item \textbf{Case IV: } `Two state' jamming. All tendons were unjammed before touchdown and jammed during midstance just before push-off.
\end{enumerate}

For each case, the walking test was repeated for four cycles to obtain the GRF-time data, as shown in Fig. \ref{fig:human} (b). The GRF data was then filtered to remove noise and segmented into individual cycles. The GRF data of each cycle was then averaged to calculate the mean resultant GRF and error bars, indicated by the shaded regions in Fig.\ref{fig:walkingresults} (a). The peak GRF after touchdown and maximum push-off force was extracted from each cycle, to produce the box-and-whiskers plot presented in Fig.\ref{fig:walkingresults} (b). For Case I, where there are no observable peaks at push-off, the maximum GRF reading obtained after the 6\,s mark was computed.

\subsection{Walking Experimental Results}

The mean resultant GRF obtained from the walking experiments are shown in Fig.\ref{fig:walkingresults} (a). Similar to the human gait, the maximum peak GRF occurs during the `loading response' phase just after touchdown and before the midstance, where the foot flattens against the ground and accepts the weight of the entire  leg. As predicted, when all tendons are unjammed (Case I) the mean touchdown force (49.4\,N) was significantly lower than the rest of the cases. Although the GRF plot showed best damping response, indicated by the gradual rise in force, the overall reduction in stiffness of the leg resulted in unpredictable motions during loading. This is indicated by the lack of a distinct touchdown and push-off peak in the GRF-time plot, which was observed in all other cases. The large variability in measured GRF between cycles was also observed, indicated by the 
large shaded error region after the 4.5\,s mark. During this time, the JEG was performing the push-off phase, and the large variability in measured GRF may be explained by the leg's inability to support its own weight, resulting in the fluctuating GRF. Furthermore, it was also observed that there was a delayed response in actuation of the leg, leading to varying foot-ground contact times, as indicated by the large shaded error region after the 8\,s mark. For some cycles, the contact duration was also found to extend beyond 10s. In all other cases (Case II, III, and IV) the foot loses contact with the ground consistently, and the resultant GRF immediately goes to zero.

In contrast, when all tendons are jammed (Case II), more distinguished touchdown and push-off peaks were observed, with a slightly higher mean resultant forces calculated as 58.8\,N and 13.6\,N respectively. The human-inspired gait (Case III) produced a closely identical force-profile, but with a slightly lower mean touchdown and push-off force peaks of similar magnitude, 55.9\,N and 12.6\,N. However, an important observation to note is that the peak forces obtained for the repeated walking cycles were significantly more consistent compared to other cases, as shown by the shorter box plot in Fig. \ref{fig:walkingresults} (b). This is a good indication of repeatability and control of the behaviour of the leg.

An initial speculation was that the JEG should perform best for Case IV, when the all the tendons were unjammed during touchdown for maximum impact absorption and later jammed for push-off for maximum thrust. However, results show that although it has a good initial damped response, as indicated by the steady increase in force after touchdown, the magnitude of the touchdown and push-off peak force is the highest compared to the other cases, 62.7\,N and 19.2\,N respectively. Similar to Case I, the results show no prominent force peaks, resulting in a large variability in both touchdown and push-off forces. The sudden spike in force seen around the 4.3\,s mark, was due to the sudden stiffening of the tendons when the vacuum is turned on. Hence, suddenly transitioning from a fully unjammed to a fully jammed state may introduce a jerk or unwanted disturbance, which may affect the stability of the system. With the human-like jamming (Case III), there was almost no observable spikes in force, despite that the pneumatic solenoid valves were rapidly switching on and off to draw air out of the tendons throughout the gait.

\begin{figure*}[!t]
\centerline{\includegraphics[width=1.9\columnwidth]{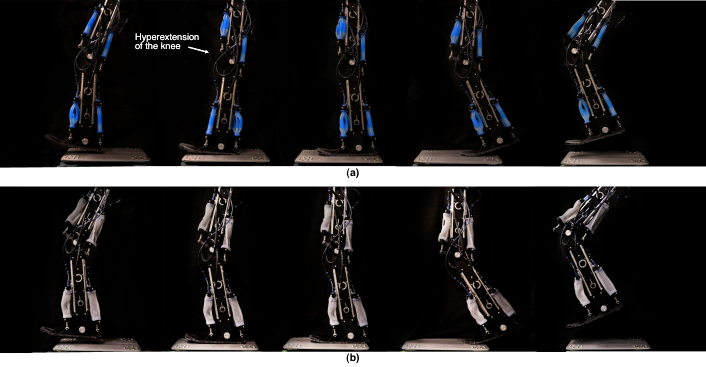}}
\caption{Walking experiments. (a) Tendons are all unjammed, with membranes removed to visualise the compression and tension of the tendons. Due to lack of stiffness of the knee flexor, the knee hyperextends and leg straightens out. (b) Tendons are jammed with a human-like muscle activation pattern. The JEG was observed to walk with a smooth gait. Videos of all jamming cases are available in supplementary materials.}
\label{fig:kinematics}
\end{figure*}

\section{Discussion}

 Early on, we presented the design of a multi-material fibre jammed tendon with tunable stiffness and damping properties. We show that by altering the geometric configuration of the tendons, such as the fibre diameter and number of fibre layers, we can achieve tunable tensile properties. The multi-material fibre jammed tendons were designed for manufacturing, to enable 3D printing to be done in a single step and without need for manual construction or assembly of the individual fibres. We presented an improved FEM numerical model from our previous presented paper \cite{pinskier2022jammkle}, which is able to predict the behaviour of the fibrous media in the tendon's unjammed and jammed states. Our model was able to accurately capture the hysteric behaviour and therefore, the damping properties of the tendon, in addition to its stiffness.
 
 To demonstrate the capabilities of the fibre jammed tendons, we explored their potential in a legged robotics application by developing a fibre jammed leg, the JEG. The JEG exploits the tensile properties of the tendons mounted in an antagonistic lever-type configuration to provide variable stiffness and damping of the knee and ankle joints. There is a large complexity in understanding the behaviour of soft jammed structures and how they can be effectively used as passive compliance mechanisms in load-bearing systems such as robotic legs. As a preliminary experiment, we developed the mock-up of a robotic leg using a simple revolute joint with pairs of tendons mounted on an antagonistic lever-type mechanism. By altering the pressure, we showed that the tendons could be used to vary the impact response of a joint.

We found that at low pressures or when the tendons are unjammed, they were effective in shock absorbance and reducing impact forces travelling up the leg. However, a joint with low stiffness and damping often resulted in large angular velocities and uncontrolled joint movement upon impact. At a multi-joint level system, we see a consequence of this behaviour in both simulation and in the walking experiment. In the multibody FEM simulation, we observed that for both toe-down and toe-up simulations, the \textit{All Unjammed} configuration performed poorly upon collision with an obstacle, resulting in high impact forces and kinetic energies from the collision and subsequently, greater rebounding of the leg from the ground. In the walking experiments, the fully unjammed tendons resulted in much lower GRF at touchdown. However, as weight of the JEG is transferred to the foot, we saw large errors regions in the GRF-time plot, as a result of the poor performance of the leg under load, resulting in unstable and unpredictable movements. To visualise the behaviour of the tendons throughout the walking gait, the changing tension and compression of the tendons are studied in Fig.\ref{fig:kinematics}. The tendon membranes were removed to enable a clear visualisation of the states of the tendons. During the midstance, it was observed that the knee joint hyperextends and the entire leg straightens due to insufficient stiffness of the knee, as seen in Fig. \ref{fig:kinematics} (a). This hinders movement of the leg, in which actuators continue to move without resulting in any transmission of motion to the joints. Instead, motion from the actuators continue to stretch the unjammed tendons until the prismatic joint limits are reached. This joint instability causes the whole length of the leg to become a large moment arm about the hip joint. This results in large undesired torques about the hip joint, and consequently led to the overloading of the motor. This was also seen when the two-state jamming pattern was used, since during the initial touchdown phase, the entire leg was unjammed as well.

On the other hand, at high pressures or when the tendons were jammed, stiffness and damping properties of the tendons were greatly increased. This is beneficial for instances where joints have to respond to large external loads, such as moderating the foot-ground impact forces at touchdown, to ensure joint stability. With reference to the results of the multibody FEM simulations, we saw that \textit{All Jammed} configuration performed the best in terms of energy storage and minimising the force of impact from the collision. Consequently, lower impact kinetic energies were seen and no rebound was seen with the toe-down collision simulation. With the walking experiments, when the leg was fully jammed or when a human-like jamming pattern was applied, the knee was able to perform forward propulsion of the leg in one smooth motion, as seen in Fig. \ref{fig:kinematics} (b). This shows the critical role of the stiffness of the knee in the transfer of energy between the ankle and hip joint \cite{torricelli2016human}. Out of the four tendon jamming patterns, the human-inspired jamming (Case III) was found to show a lower magnitude in touchdown GRF compared to when the JEG was fully jammed throughout the entire cycle (Case II) or when jamming was applied after the midstance (Case IV). Results obtained were also more consistent and repeatable compared with the rest of the cases.

From the multibody dynamics FEA simulation of the JEG, we observed that jamming only the ankle tendons were more favourable than only the knee tendons, in reducing kinetic energy and impact force at touchdown. Hence, the relative force contribution of the individual tendons to the overall stiffness of the leg was investigated through the rotational perturbation experiments. Both ankle tendons were found to be the main contributors towards opposing rotational perturbations. This is consistent with previous findings in biomechanical research, where the ankle muscles were found to be the primary shock absorber and had the highest activation rate when perturbed in the anteroposterior direction \cite{shen2022neuromechanical,zelik2012mechanical,zhu2022does}. In addition to the ankle tendons, the knee extensor and flexor tendons also contribute to increase the rigidity of the leg as well, thereby effectively decelerating knee flexion during the leg loading phase. This study has provided valuable insights into the distinct roles each tendon plays in disturbance rejection, highlighting their unequal force contributions. This understanding of biologically inspired compliant mechanisms are crucial, as it paves way for more morphologically intelligent designs, where these mechanisms could be optimally tuned for different joints.

\section{Conclusion}



The study of soft robotics was originally stemmed from the inspiration to mimic the soft and compliant characteristics found in invertebrates. Vertebrates, on the other hand, possess rigid skeletons that are responsible for providing body weight support and facilitate the transmission of forces during movement. In humans, locomotion is only possible through the coordinated action of muscles and tendons, which perform majority of the work. Our research draws inspiration from the human musculoskeletal system, wherein the inherent mechanical properties of muscles are used to enable automatic stabilisation of movements in response to unexpected perturbations. This remarkable capability had motivated our research into developing a compliant articulated soft robotic leg, the JEG, which utilises multi-material fibre jammed tendons as passive compliance mechanisms to act as buffers between the actuators and belt transmission mechanisms from external loads and perturbations. The performance of the proposed tendons and the JEG was evaluated through extensive simulations and physical experiments. The main implications of our research are as follows: 
\begin{enumerate}
    \item We presented the design of a multi-material fibre jammed tendon, which can be 3D printed in one-go without need for assembly. These fibre jammed tendons offer both stiffness and damping tunability, providing researchers the freedom to customise the mechanical properties of these structures towards a specific function or application.
    \item We developed a numerical FEM model capable of accurately capturing the tendon's stiffness and damping properties. The FEM numerical model presented in the paper aims to act as a tool to facilitate rapid design exploration of these structures for diverse applications, such as legged locomotion.
    \item We demonstrate the implementation of fibre jammed structures in a dynamic load bearing application by developing the JEG, with fibre jammed tendons were placed in an antagonistic configuration on the knee and ankle, allowing improved versatility and adaptation towards uncertain environments. 
    \item  We identified the role and force contribution of each tendon pairs towards the overall stiffness of the leg. Results of the perturbation test indicated that the knee extensors, knee flexors, ankle dorsiflexors and ankle plantarflexor tendons individually play different roles in both stability and locomotion. Depending on the nature of the perturbation, specifically the direction of rotation in which the JEG was loaded, jamming specific tendons could yield vastly different dynamic effects. 
    \item  We investigated the behaviour of these compliant tendons and the walking performance of the JEG, by evaluating the changes in ground reaction forces in response to varying jammed states of the tendons. From this experiment, we were able to identify notable gait characteristics from the force results to inform of the stability of the leg.

\end{enumerate}

In conclusion, our research delves into the exploitation of fibre jammed tendons in legged locomotion. The utilisation of these tunable passive compliance structures open new doors to morphologically intelligent designs that are versatile and capable of responding to external perturbations during locomotion with minimal active control. By using tunable jammed structures, we unlock a promising avenue for the development of more cost-effective and lightweight robotic legs. This breakthrough holds significant potential in revolutionising the field of robotics, offering unprecedented capabilities for robust and efficient locomotion in unstructured environments.


\section{Future Work}
This study focuses primarily on the leg's response towards rotational perturbations. However, it is crucial to acknowledge that perturbations can manifest in many forms, such as translational horizontal perturbations (e.g the sudden stop of a moving bus) or vertical perturbations (e.g missing a step or jumping from an elevated surface). However, given the substantial disparity between these perturbation types, simulating them would require distinct experimental setups. There are numerous opportunities in the future to investigate the leg's response to diverse loading scenarios, to improve our understanding on how soft jammed structures could be effectively implemented in the joints of legged systems. 

In biological systems, stability is achieved by the properly tuned material properties of the musculoskeletal system, as well as through neural reflexes and higher brain control. This paper does not address the implementation of an active control system with feedback to dynamically adjust the length of the tendons or impedance of the system. To address this gap in research, a future study may include the implementation robotic vision or sensorising the JEG, and using machine learning techniques to classify the nature of the terrain or footholds before the foot makes impact. A controller could then be effectively employed to preemptively alter the properties of the tendons in response to predicted perturbations, similarly how humans use visual cues and prior knowledge of the environment to execute anticipatory strategies, and activating their muscles before encountering perturbations.

\section{Acknowledgements}
The authors would like to thank Tom Molnar for providing support in the initial development of the walking controller and Sarah Baldwin for help with 3D printing of the tendons.

\bibliographystyle{IEEEtran}

\newpage
\vspace{-33pt}
\begin{IEEEbiography}[{\includegraphics[width=1in,height=1in,clip,keepaspectratio]{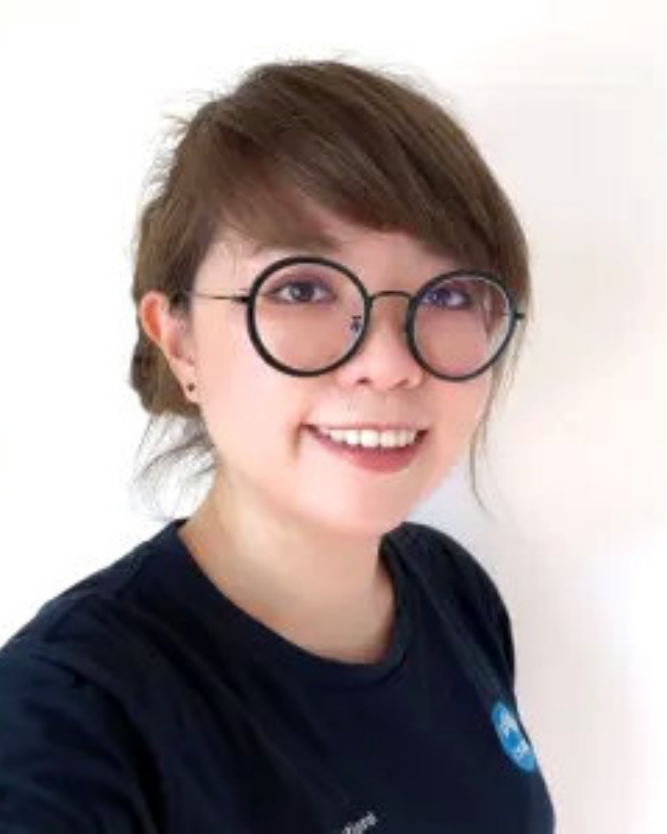}}]{Lois Liow} is a Robotics Engineer at CSIRO Data61. She graduated from Imperial College London in Design Engineering with an integrated Masters of Engineering (MEng) in 2019. During her time at Imperial College London, she conducted research in robotic manipulation and in the design of underactuated tendon-driven robotic hands. Since joining CSIRO, she has been providing engineering support to the team and has also been performing research in articulated soft robotic systems, as well as space manipulation technologies.
\end{IEEEbiography}

\begin{IEEEbiography}[{\includegraphics[width=1in,height=1in,clip,keepaspectratio]{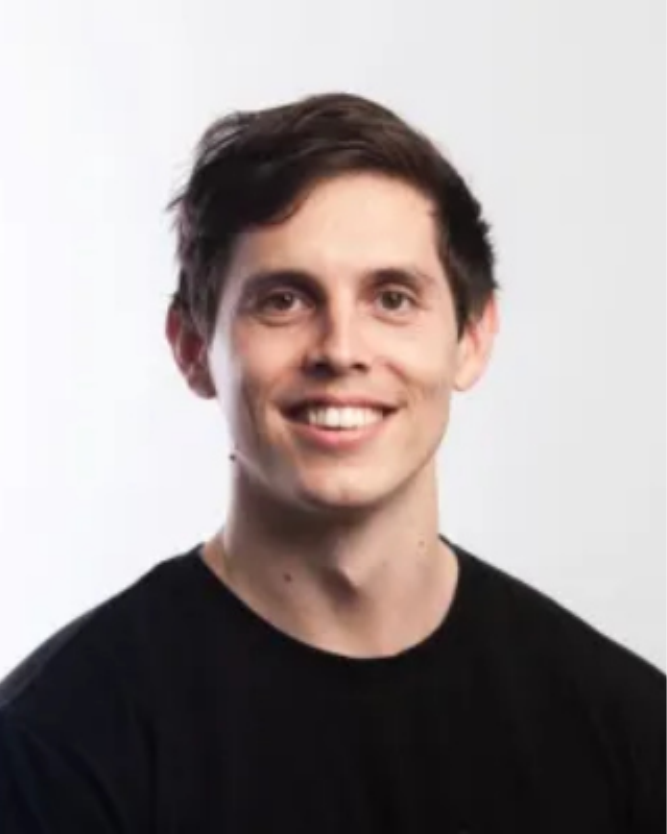}}]{James Brett} is a Senior Mechatronic Engineer who graduated Griffith University in 2014. After graduating, he began work as a Design Engineer within the Industrial Automation and Robotics field. James commenced work with CSIRO in 2016 and he currently works as a Mechatronics Engineer in the Robotics and Autonomous Systems Group at CSIRO Data61.
\end{IEEEbiography}

\begin{IEEEbiography}[{\includegraphics[width=1in,height=1.in,clip,keepaspectratio]{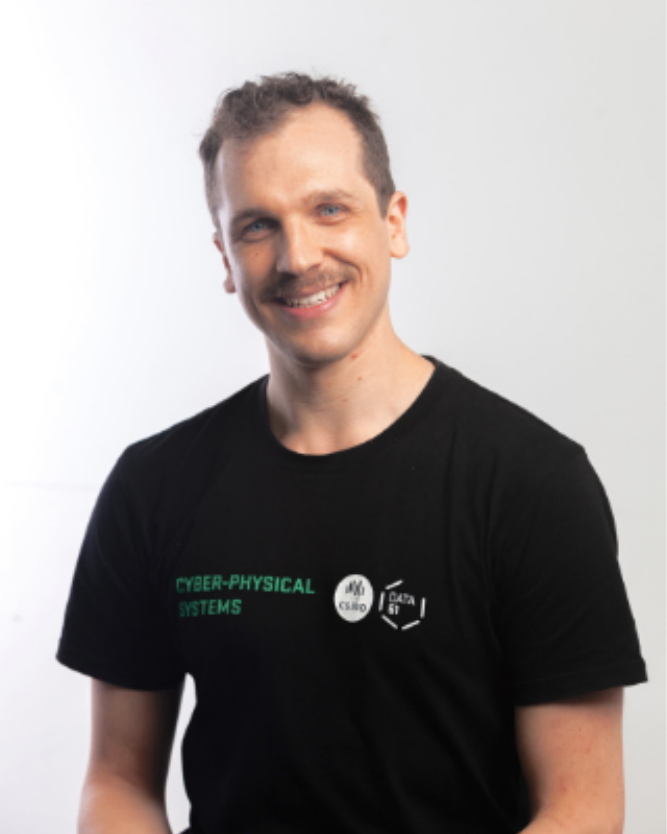}}]{Joshua Pinskier}
is a Postdoctoral Research Fellow at CSIRO Data61. He completed dual bachelors in Mechatronics Engineering and Commerce at Monash University in 2015, before completing his PhD in Mechanical Engineering at Monash in 2019. His PhD research into design optimisation of compliant mechanisms for haptic-guided nanomanipulation received multiple awards including Best Paper at 3M-Nano 2015. His current research explores topology optimisation methods to explore complex design spaces and generate high performing soft robotic components.
\end{IEEEbiography}

\begin{IEEEbiography}[{\includegraphics[width=1in,height=1.in,clip,keepaspectratio]{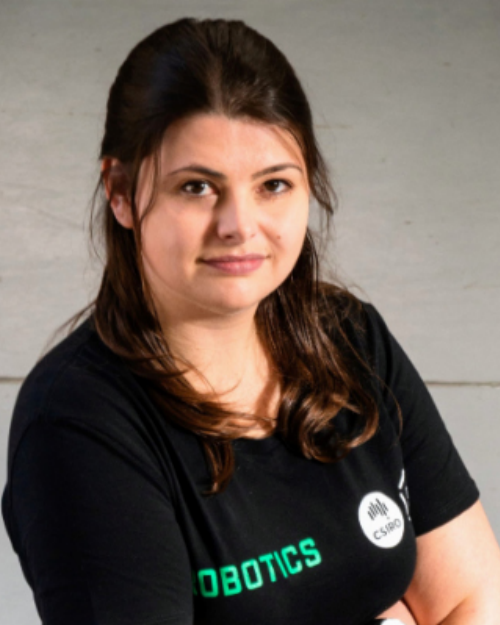}}]{Lauren Hanson}
Lauren is a Senior Mechanical Engineer in the Robotics and Autonomous Systems Group at CSIRO. She studied a dual degree of Mechanical Engineering (Honours) and Science at Monash University. Since joining CSIRO she has worked on multiple robotics projects, including leading development of mechanical systems for the CSIRO Data61 DARPA Subterranean Challenge team and being an active member of the soft robotics research team. She is currently supporting the CSIRO space research efforts, and is designing hardware for a research payload to be launched to the ISS later this year.
\end{IEEEbiography}

\begin{IEEEbiography}[{\includegraphics[width=1in,height=1.in,clip,keepaspectratio]{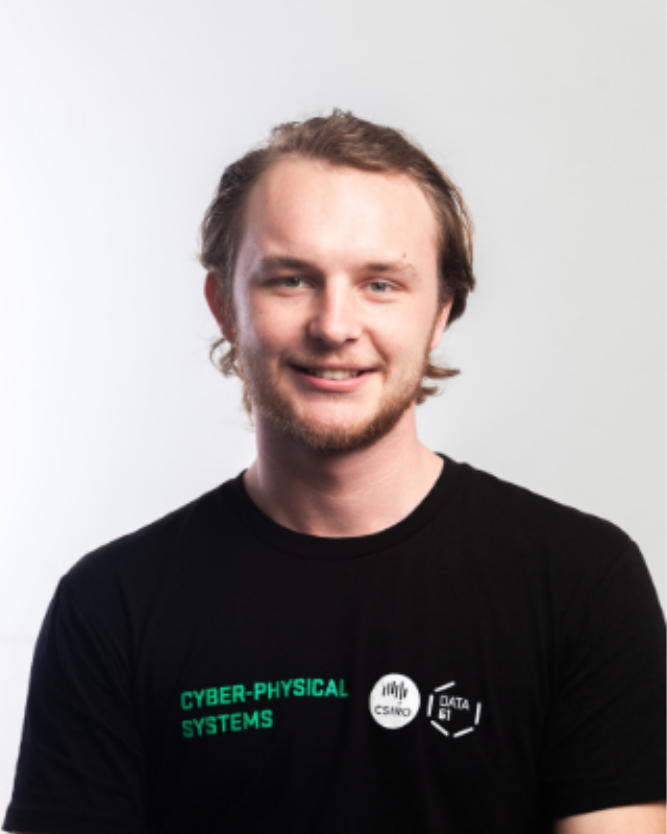}}]{Louis Tidswell} started with CSIRO in November 2021 to complete his engineering internship with Data61. He then completed his Mechatronics Engineering honours project with CSIRO and Queensland University of Technology during 2022 in the field of soft robotics. After graduating, he accepted a job as a graduate Mechatronics Engineer with Athena Artificial Intelligence.   
\end{IEEEbiography}

\begin{IEEEbiography}[{\includegraphics[width=1in,height=1.in,clip,keepaspectratio]{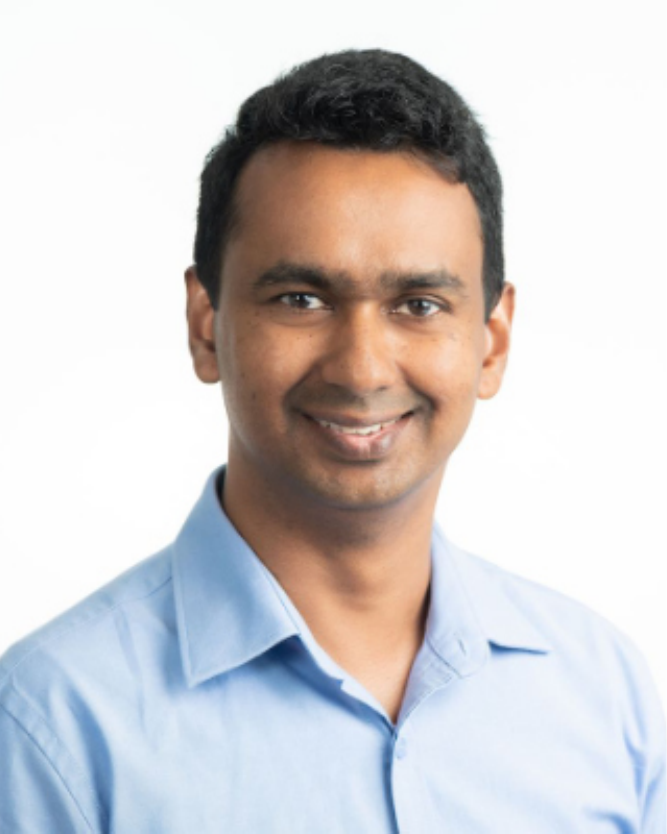}}]{Navinda Kottege}
 (M'05-SM'14) received his BSc in Engineering Physics (University of Colombo) in 2003 and his PhD in Engineering (ANU) in 2009. He is the Research Director responsible for Robotics, Computer Vision and Distributed Sensing at CSIRO’s Data61. Navinda initiated and led legged robot research within CSIRO since 2011, with a focus on navigation in unstructured environments. Navinda is a senior member of the IEEE, and a former Chair of the IEEE Queensland joint chapter for Control Systems/Robotics and Automation Societies. He is an Adjunct Associate Professor at both Queensland University of Technology and University of Queensland.
\end{IEEEbiography}

\begin{IEEEbiography}[{\includegraphics[width=1in,height=1.in,clip,keepaspectratio]{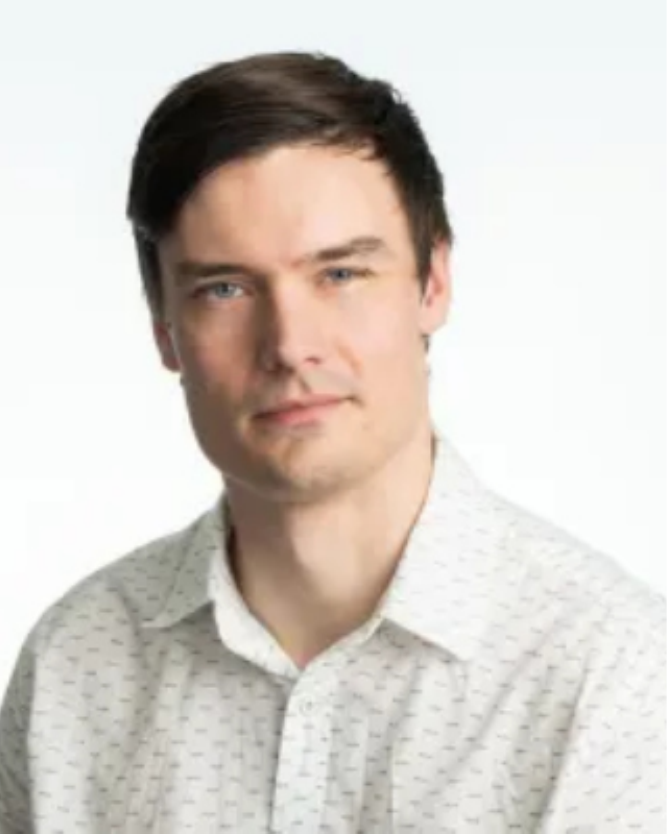}}]{David Howard} David is a Principal Research Scientist and Soft Robotics Research Leader at CSIRO, Australia’s national science body. He received his BSc in Computing from the University of Leeds in 2005, and the MSc in Cognitive Systems at the same institution in 2006. In 2011 he received his PhD from the University of the West of England. He is a member of the IEEE and ACM, and an avid proponent of education, STEM, and outreach activities. His work has been published in IEEE and Nature journals. His interests include nature-inspired algorithms, learning, soft robotics, the reality gap, and evolution of form. 
\end{IEEEbiography}

\vspace{400pt}

\end{document}